\documentclass[lettersize,journal]{IEEEtran}
\usepackage{amsmath,amsfonts}
\usepackage{algorithm}
\usepackage{algorithmicx}
\usepackage{algpseudocode}
\usepackage{array}
\usepackage[caption=false,font=normalsize,labelfont=sf,textfont=sf]{subfig}
\usepackage{textcomp}
\usepackage{stfloats}
\usepackage{url}
\usepackage{verbatim}
\usepackage{graphicx}
\def\BibTeX{{\rm B\kern-.05em{\sc i\kern-.025em b}\kern-.08em
    T\kern-.1667em\lower.7ex\hbox{E}\kern-.125emX}}
\usepackage{balance}

\usepackage[numbers,sort&compress]{natbib}

\makeatletter
\def\thanks#1{\protected@xdef\@thanks{\@thanks
        \protect\footnotetext{#1}}}
\makeatother

\begin{document}

\title{PLD-SLAM: A Real-Time Visual SLAM Using
Points and Line Segments in Dynamic Scenes}


\author{Baosheng Zhang}

\maketitle

\begin{abstract}
    In this paper, we consider the problems in the practical application of visual simultaneous localization and mapping (SLAM). With the popularization and application of the technology in wide scope, the practicability of SLAM system has become a new hot topic after the accuracy and robustness, e.g., how to keep the stability of the system and achieve accurate pose estimation in the low-texture and dynamic environment, and how to improve the universality and real-time performance of the system in the real scenes, etc. This paper proposes a real-time stereo indirect visual SLAM system, PLD-SLAM, which combines point and line features, and avoid the impact of dynamic objects in highly dynamic environments. We also present a novel global gray similarity (GGS) algorithm to achieve reasonable keyframe selection and efficient loop closure detection (LCD). Benefiting from the GGS, PLD-SLAM can realize real-time accurate pose estimation in most real scenes without pre-training and loading a huge feature dictionary model.
To verify the performance of the proposed system, we compare it with existing state-of-the-art (SOTA) methods on the public datasets KITTI, EuRoC MAV, and the indoor stereo datasets provided by us, etc. The experiments show that the PLD-SLAM has better real-time performance while ensuring stability and accuracy in most scenarios. In addition, through the analysis of the experimental results of the GGS, we can find it has excellent performance in the keyframe selection and LCD.
\end{abstract}


\section{Introduction}
\IEEEPARstart{S}{imultaneous} Localization and Mapping (SLAM) is a technology to estimate the state of the robots and build surrounding maps in unknown environments \cite{cadena2016past}. In the past few decades, outstanding achievement has been made in SLAM and competent for most of the demanding tasks, such as autonomous driving \cite{geiger2012we}, underwater exploration \cite{kinsey2006survey}, and micro aerial vehicle (MAV) navigation \cite{faessler2016autonomous}. The mass applications of visual SLAM have further proved the excellent performance of SLAM in location and building maps and promoted the further development of related technologies.

Currently, the most of SLAM systems based on Lidar \cite{zhang2017low}\cite{huang2021disco} or camera \cite{durrant2006simultaneous, 2017VINS, pumarola2017pl}. Since Lidar can obtain highly accurate point clouds in a wide range around and is robust to the wide range of lighting changes, the SLAM based on 2D or 3D Lidar once became the mainstream of SLAM systems \cite{huang2021disco}. However, with improvements in the device performance and the massive emergence of vision algorithms in recent years, a large number of visual SLAM approaches based on various visual sensors have been proposed. Visual SLAM has played a more important role in research and application in virtue of its information-rich measurements, low cost, and low energy consumption. With the popularity of visual odometry (VO) and visual SLAM technology, the more practical problems in the real scenes emerge gradually and become the new popular research topics. For the benefit of research, we summarize the new primary requirements for the visual SLAM systems as follows:
\begin{enumerate}
    \item Improving the capability of the system to cope with the dynamic environment;
    \item Obtaining the robustness and accuracy results while ensuring the real-time performance of the system;
    \item Improving the universality of the system, e.g., implementing the system immediately without additional preprocessing in the new different environments.
\end{enumerate}
In this paper, we consider the above-mentioned challenges and proposed effective solutions.

Finding feature correspondences between two different images is a necessary task in feature-based indirect visual SLAM \cite{sun2019guide}\cite{fu2021fast}. To deal with the poor textured environment, we introduce line segments as basic features into our system. For line features, we use the LSD detector \cite{von2008lsd} to extract salient lines in the image like most line feature-based approaches \cite{pumarola2017pl}\cite{gomez2019pl}\cite{zuo2017robust}. In the subsequent feature matching process, keeping the number and correct of matches are the most concerning issues and researchers have contributed meaningful work in this field \cite{ma2015non, ma2014robust, yi2018learning}. However, these feature mismatch removal methods suffer from the drawbacks time-consuming and the limitation of the feature type. In this article, we respectively propose two novel local feature outlier filters for point and line features that address these problems and aim to remove greatly mismatched features one time. Filtering false line feature matched is not a trivial task, the key challenge is that the line features involve a large range in view that cannot be divided locally by regular grids like the point features. In this work, we tackle the issues using a purely visual-based approach. We focus on the line feature itself and construct the circular domain is parameterized by the mid-point and length of each line. Then creating a local line feature domain refer to the circular domain relationship between the line features, removing the outlier matching lines based on the position and direction of all the line features in the line domain. Further discussion about the feature mismatching will be formally addressed in section \ref{sec:Track_a} and \ref{sec:Track_b}.

Among the existing SOTA visual SLAM and VO approaches, such as ORB\_SLAM \cite{mur2015orb} and SVO \cite{forster2016svo} have estimated the accurate state of robots by optimizing the feature re-projection errors. However, these indirect or semi-direct methods suffer from drawbacks arising from the assumption of static environments. The dynamic features attached to dynamic objects as the critical challenge for conventional visual SLAM, which greatly reduces the location and map reconstruction accuracy. The idea to solve the highly dynamic environment problem is to accurately detect dynamic regions in the view and remove outliers, then the system achieves the motion estimation between different frames based on the remained static features. To retain enough stable features, we refer to our previous work \cite{Ma2022DynPLSVOAN}, using the $dynamic$ $grid$ algorithm which utilizes the motion model between adjacent frames to sign and remove dynamic point features to solve the problem of dynamic objects in real-world scenes. In addition, we extended the dynamic point detection algorithm to the line features, a brief conclusion and discussion are given in section \ref{sec:Track_c}

Besides, loop closure detection (LCD) is an essential element for visual SLAM working in long-term and large-scale scenes, and it is increasingly a concern by researchers. An excellent LCD requires the capacity to recognize a previously visited place from current camera measurements and reduces the drift accumulated of all keyframes in the loop \cite{angeli2008fast}\cite{cattaneo2022lcdnet}. In exiting popular SOTA visual SLAM, the LCD proposed is almost implemented by the feature-based bag-of-words (BoW) \cite{galvez2012bags} model or its variants. According to the different times of building BoW models, the BoW-based LCD is divided into static and dynamic models. The off-line feature dictionary model needs to build a BoW model for all field features in the dataset before the system operation, and loading the huge model at the initial stage of system implementation, which is not conducive to the universality of the system to the new environment nor the real-time performance in the automatic implementation process. On the contrary, the dynamic BoW model creates a feature dictionary for each frame during the system execution to realize LCD in subsequent frames. However, although the BoW model can achieve high accuracy performance with a huge visual dictionary, its unable to meet both the real-time and universality (requiring pre-training vocabulary model for new scenes) of the system.

In this work, we present a novel image similarity moble based on global gray similarity (GGS), which extracts the field of view information through the fast global gray value distribution, and realizes the similarity evaluation between different images (see section \ref{sec:Track_e}). The function of the GGS algorithm introduced into the system is reflected in two parts. First, the LCD in PLD-SLAM based on the GGS achieves SOTA accuracy without additional pre-training models and greatly improves the adaptability of the system to the new environment and the real-time performance of the system. In particular, the GGS-based LCD also has an extremely important reference value for the LCD works of conventional directly or LiDAR SLAM, etc., which without feature detection. Second, a keyframe selection based on the GGS proposed in this work, optimally selects reasonable keyframes for the balance of accuracy and efficiency of the system.

The main contributions of this article can be summarized as follows:
\begin{itemize}
    \item We present a complete stereo visual SLAM system based on point and line features in dynamic scenes, coined PLD-SLAM, which achieve SOTA accurate result. This system has real-time performance and better adaptability compared with traditional visual SLAM;
    \item Two effective purely visual feature mismatching filter algorithms to remove outliers of point and line features, which balances the feature tracking needs between the accuracy and computational complexity;
    \item A novel keyframe selection refers to GGS, which meets the requirements of LCD and improves the real-time of the system.
    \item A light-weight and fast LCD based on the GGS of keyframes;
\end{itemize}

The rest of our paper is organized as follows. After a review of the relevant background literature in Section II, the main details proposed in the paper are described in section III. Subsequently, Section IV provides qualitative and quantitative results of the performance of PLD-SLAM both on the KITTI dataset [9] and in a real-world environment to demonstrate the effectiveness and accuracy of the system. Finally, a brief conclusion and discussion are given in section V.

\section{Related Work}
\subsection{Visual SLAM}
Based on the type of sensors, most visual SLAM and VO systems can be divided into Monocular \cite{2017VINS} \cite{pumarola2017pl}, Stereo \cite{gomez2019pl}, and RGB-D \cite{kerl2013dense} method. PTAM \cite{klein2007parallel} is one of the keyframe-based monocular framework visual SLAM, which first introduces the idea of feature tracking and mapping into parallel threads. As a SOTA SLAM, ORB\_SLAM \cite{mur2015orb} begins with PTAM, employing local Bundle Adjustment (BA), Pose Graph Optimization (PGO), and LCD to improve the accuracy and robustness of the system. Furthermore, the method in the \cite{pumarola2017pl}, simultaneously leverages points and lines information to deal with the poorly textured environment. However, all of these monocular methods have a critical drawback in the initialization and estimation accuracy. The RGB-D scheme can address the above issues and directly build dense maps via the depth information of each pixel. Since the application range is a key limitation, such as outdoor and underwater, RGB-D SLAM is also not able to applicate in wide scope. 

To overcome these challenges, various stereo methods solutions have been proposed. In the past decades, the stereo framework in ORB\_SLAM2\cite{mur2017orb} was one of the SOTA stereo frameworks, it used ORB as the point feature and employ BA to eliminate the drift of trajectory estimation, balancing the real-time and accuracy of the system. In addition, Y Zhou $et$ $al.$ present a parallel tracking-and-mapping approach system based on the stereo event cameras \cite{zhou2018semi}, it tackles challenging scenarios in robotics, such as low-light, high-speed, and high dynamic range scenes while running in real-time.

\subsection{Feature-based SLAM}
\begin{figure}[!t]
    \centering
    \subfloat{
        \includegraphics[width=3.3in]{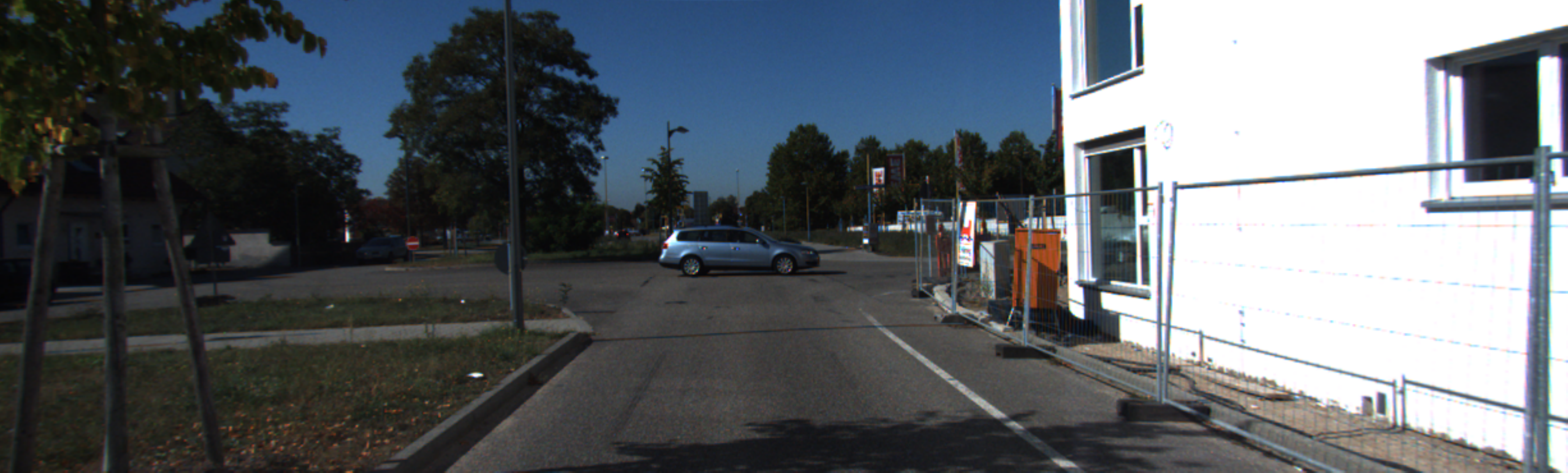}
    }
    
    \subfloat{
        \includegraphics[width=3.3in]{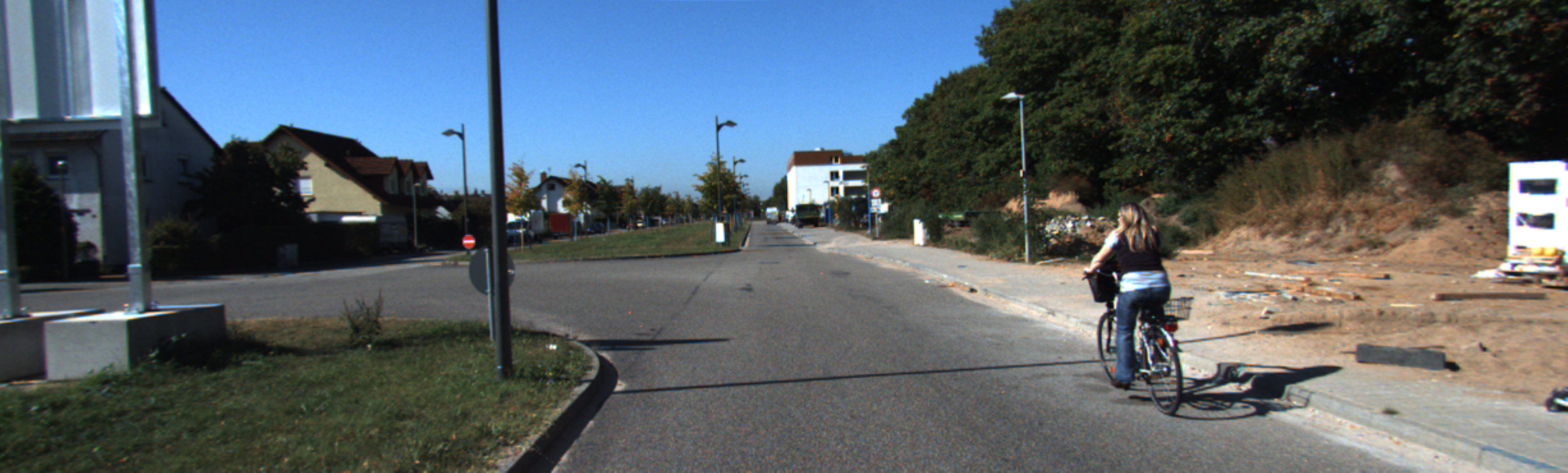}
    }
    \caption{The moving car and pedestrian in the KITTI-06 dataset.}
    \label{fig:rel1}
\end{figure}
Among the above approaches almost the indirect method, currently exiting visual SLAM can be divided into direct \cite{engel2014lsd}\cite{engel2017direct} and indirect (feature-based)\cite{pumarola2017pl}\cite{mur2015orb}\cite{gomez2019pl} according to visual measurement processing. Compared to the feature-based methods, the direct method directly uses pixel intensity to compute the camera motion by minimizing photometric errors, instead of detecting and tracking features. However, they are greatly sensitive to brightness changes and constrained to narrow baseline motions \cite{javidnia2017accurate}, most of the researchers used feature-based ones instead due to their robustness and accuracy of estimation. Furthermore, most the feature-based SLAM, no matter monocular or stereo, use point features as their common visual feature and compute corresponding 3D points in the camera frame by correct matching between the detected point features. Among these algorithms, we focus on PMHT SLAM \cite{davey2007simultaneous}, SVO \cite{forster2016svo}, and the ORB\_SLAM2 presented by R Mur-Artal $et$ $al.$ \cite{mur2017orb}, is a typical point feature-based visual SLAM, which merge multiple frameworks for various vision sensors. It is based on the fast and convenient handle ORB features to recover trajectory and build sparse point cloud maps. 

However, all these approaches with only point features can obtain robust and accurate motion estimation in rich-texture scenes. When facing scenes with poor information, the motion estimation accuracy could degrade significantly. To achieve more robust and better accuracy performance, researchers introduced line features to improve feature detection capability in the low-texture environment \cite{zuo2017robust}\cite{zhang2015building}\cite{gomez2016pl}. The method \cite{zhang2015building} as one of the line-based excellent methods, presents a graph-based stereo visual SLAM system using straight lines as only visual features. In this work, using two different representations to parameterize 3-D lines to meet the needs of projection and optimization, the same strategy can be found in \cite{zuo2017robust}. Other systems utilize other features and line features at the same time, as described in \cite{pumarola2017pl} and \cite{gomez2019pl}, the point as the most mature and convenient visual feature, complementing the advantages of line features and providing more structural information from the environment over only one kind of feature map. However, no matter direct or feature-based approaches, all existing traditional visual SLAM approaches suffer from drawbacks arising from the dynamic objects in the real scenes, such as moving cars and pedestrians shown in figure \ref{fig:rel1}. In the following, we review the state of the art of visual SLAM or VO systems for highly dynamic environments.

\subsection{Dynamic SLAM}
To eliminate the impact of the dynamic environments, there are several types of moving object detection algorithms that extract feature points and clustered features to distinguish the background and the foreground, in this way to estimate the motion of the camera \cite{viswanath2015background}\cite{kim2020moving}. In the \cite{minaeian2018effective}, S Minaeian $et$ $al.$ proposed an effective, efficient, and robust method to accurately detect and segment multiple independently moving foreground targets. Then tracking background key points using pyramidal and using a sliding window for detecting multiple moving targets. Other methods with depth information can be applied, the method \cite{du2020accurate} provides a more accurate dynamic 3D landmark detection method, and first introduce an efficient initial camera pose estimation method based on distinguishing dynamic from static points using graph-cut RANSAC. Recently, learning-based computer visual methods are massive emerging, researchers utilize various deep learning and convolutional neural networks (CNN)-based methods to solve the dynamic object \cite{bilen2016weakly}\cite{wang2017joint}. It is noted that, in general, all these deep learning methods are likely to be computationally expensive and time-consuming, which is not feasible for system applicate in the real scene.
Currently, the maturity of SLAM as a research field has motivated the practicability of SLAM systems. In this paper, we proposed a real-time visual SLAM in dynamic scenes, the several novel strategies have reference value for SLAM system real application.

\section{Technical Approach}
\subsection{System Overview}
\begin{figure*}[!t]
    \centering
    \includegraphics[width=\textwidth]{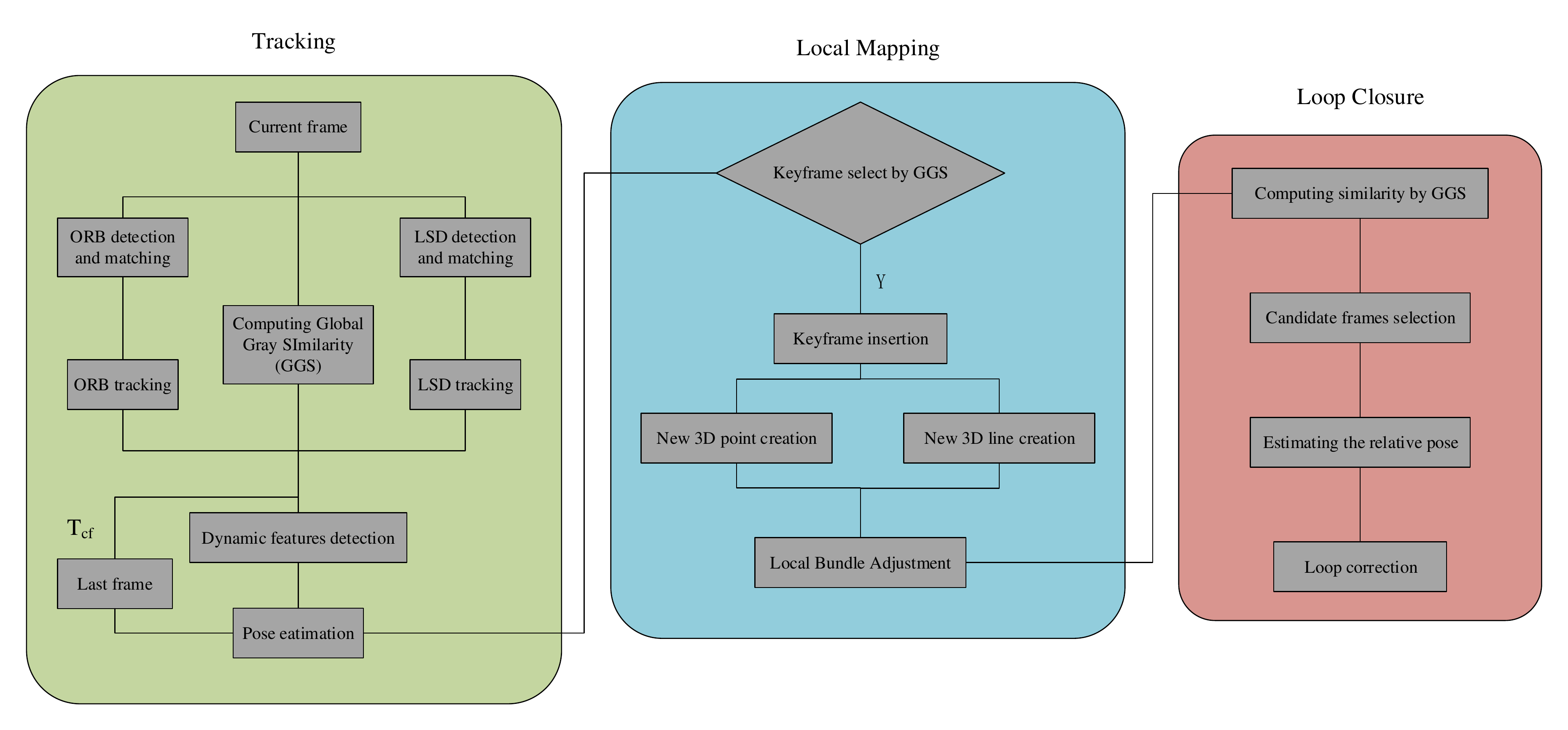}
    \caption{System overview. PLD-SLAM involved three main threads: tracking, local mapping, and loop closure. The feature detection and GGS computation of new frames parallel in the tracking thread. The local mapping achieve when a keyframe was selected, and the loop closure behavior when the correct loop closure was detected.}
    \label{fig:SysO_1}
\end{figure*}

The system overview of our approach is given in figure \ref{fig:SysO_1}, following most feature-based SLAM frameworks, we divide PLD-SLAM into the three components: Tracking, Local Mapping, and Loop Closure detection. We next briefly review the pipeline of the work.

First, in the tracking components, we aim to estimate the pose of the new frame and preparation for the operation of the system. In the initial stage of the component, we compute the GGS of the image, which is the key parameter of the keyframe selection and loop closure detection. Notice that, normally, selecting the left view to participate in the later work on stereo visual SLAM. To reduce system time consumption, we only obtain the GGS of the left view in the new frame. Meanwhile, we detect ORB and LSD and match the corresponding features to each other to recover 3D points and lines as our preview work DynPL-SVO. Subsequently, we utilized two effective mismatching filters for points and lines, respectively, to detect and remove the false feature matching. The details are described in section \ref{sec:Track_a} and \ref{sec:Track_b}. Finally, the matched point and line features on the current frame are tracked with corresponding features on the previous frame and computing the camera motion by minimizing feature re-projection errors. However, the dynamic features in the field of the view reduce the accuracy and robustness of the system. To overcome the dynamic environment, we based on the $dynamic$ $grid$ algorithm present by our preview work \cite{Ma2022DynPLSVOAN}, and extended the dynamic point detecting algorithm to the line features. So far, we initial the camera pose estimation of the new frame. Further discussion about the feature tracking procedure will be formally addressed in section \ref{sec:Track}. The next components are keyframe insertion and updating the local map.

Before updating the local map, we need to determine when the new frame meets the conditions and is selected as the keyframe. In this study, we employ the GGS above-mentioned as the core to detect the keyframe, then insert the keyframe selected to the local map. The quality of the map depends on the accuracy of the keyframe pose and landmark state. Thus, we match the feature on the new keyframe with the feature on the preview keyframe and the local map. Next, use a co-visibility graph between local keyframes and the feature positions observed by them to build a local BA model to correct elements in the local map. A detailed description of this procedure will be presented in section \ref{sec:LM}.

During the local mapping, loop closure detection utilized GGS to calculate the similarity between each inserted new keyframe and all existing keyframes, as described in section \ref{sec:LC_1}. Then, select the correct looped keyframes which meet the two stages conditions, and correct all the keyframes poses involved in the loop through a pose-graph optimization (PGO).

Finally, we achieved global optimization via a global BA optimization identical to the ones presented in ORB-SLAM.

\subsection{Tracking}
\label{sec:Track}
For each input stereo frame, we implement the necessary initialization to obtain its rough relative pose transformation and the observed feature positions in the world. Similar to most traditional feature-based SLAM methods, in this paper, we transform the frame pose estimation into solving a nonlinear least square problem. However, these approaches do not work well in dynamic environments. This section reviews the most important aspects of our previous work \cite{Ma2022DynPLSVOAN}, which deals with the visual odometry estimation in the dynamic scenes, and also with the GGS strategy.

\subsubsection{Point feature detection and tracking}
\label{sec:Track_a}
\begin{figure}[!t]
    \centering
    \includegraphics[width=3.4in]{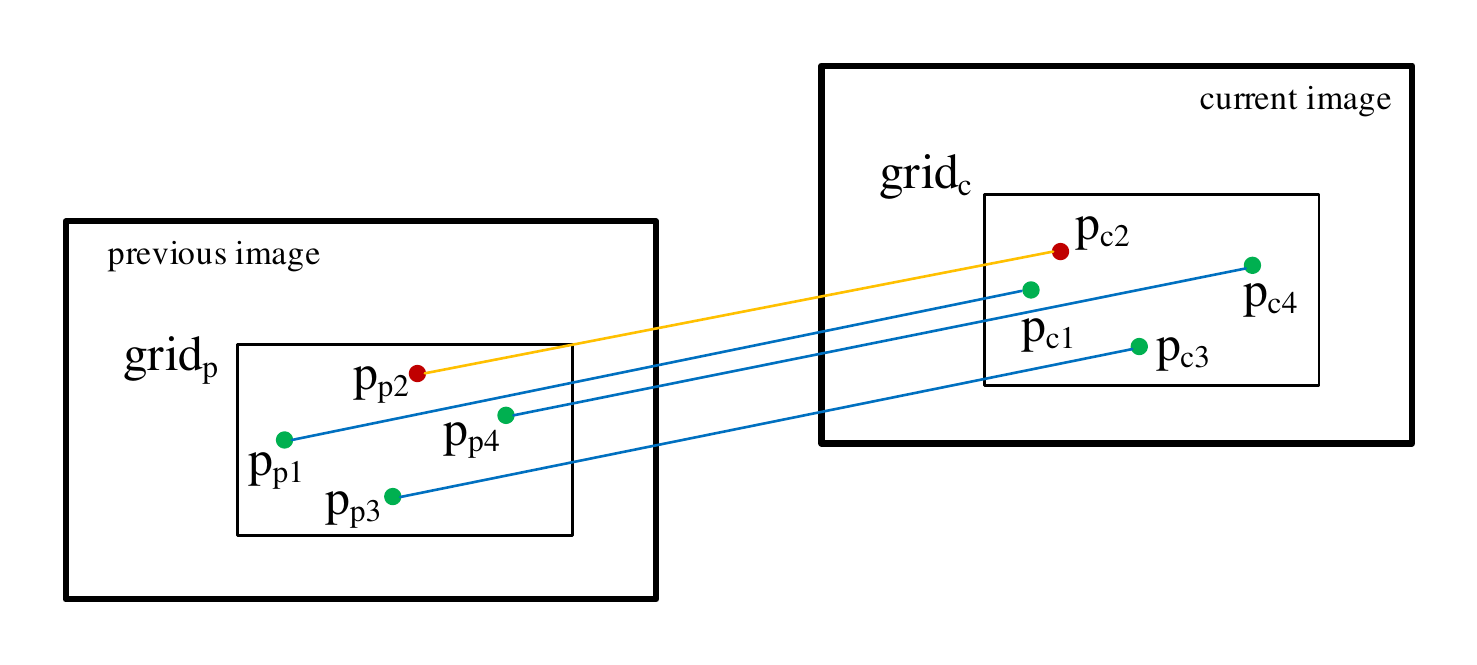}
    \caption{The point feature tracking process between the adjacent frames. We obtain the rough point features tracking ($p_{p1,2,3,4}$ - $p_{c1,2,3,4}$) between the two different frames based on point features description, and they are located in two corresponding grids, $grid_p$ and $grid_c$, respectively. Among these tracked point features, the mis-tracking $p_{p2}$ - $p_{p2}$ with an abnormal $grid_{cval}$ compare with other correct point feature trackings.}
    \label{fig:Track_1}
\end{figure}
For point feature, we use the ORB algorithm, since its balance between performance and efficiency meet the need of the system. To ensure the low time consumption, we cited the local matching scheme used in \cite{Ma2022DynPLSVOAN}. In this way, the matching point features between the left and right view in the corresponding local origin greatly reduce the number of candidates. Furthermore, to remove the point feature mis-tracking between two different frames, we advance an effective point feature outliers filter, which constraint relationship between the target point feature and other point features in the same grid to filter out outliers (see figure \ref{fig:Track_1}). First, the meshing of all detected point features, i.e., dividing the image evenly into 64×48 grids and storing all point features according to their grid positions in the image. We obtain the rough point features matching ($p_{p1,2,3,4}$ - $p_{c1,2,3,4}$) between the previous frame  and the current frame. Considering the motion model of previous frame, the correctly matched point features $p_p$ and $p_c$ have similar relative relationships with other surrounding feature points. Based on this premise, computing meaning cross between the target point location and the coordinate of all other points in the same grid, named $grid_{cval}$, as shown following:

\begin{equation}
    \begin{aligned}
      g_{lk} = \frac{1}{n}\sum_{i=1}^n \begin{bmatrix}x_{lk} \\ y_{lk}\end{bmatrix} \times \begin{bmatrix}x_{li} \\ y_{li}\end{bmatrix}\\
      g_{rk} = \frac{1}{n}\sum_{i=1}^n \begin{bmatrix}x_{rk} \\ y_{rk}\end{bmatrix} \times \begin{bmatrix}x_{ri} \\ y_{ri}\end{bmatrix}
    \end{aligned}
\end{equation}
where $g_{lk}$ and $g_{rk}$ refer to the kth matched point feature pair $grid_{cval}$ in the left and right image, respectively, and $n$ is the number of matched points in the grid. In this way, we can find that the $grid_{cval}$ between the mismatching $p_{r2}$ and $p_{l2}$ is significantly different from other correct matching, i.e., 
\begin{equation}
    |g_{l2} - g_{r2}| > \alpha (g_{lm} - g_{rm})
\end{equation}
where $g_{lm}$ and $g_{rm}$ represent the meaning $grid_{cval}$ in the $grid_l$ and $grid_r$, respectively, and $\alpha$ has been set to 1.5 in our experiments. Finally, we repeat the above operation for all matched point features and remove the outlier in each grid.

\subsubsection{Line feature detection and tracking}
\label{sec:Track_b}

\begin{figure}[!t]
    \centering
    \includegraphics[width=3.4in]{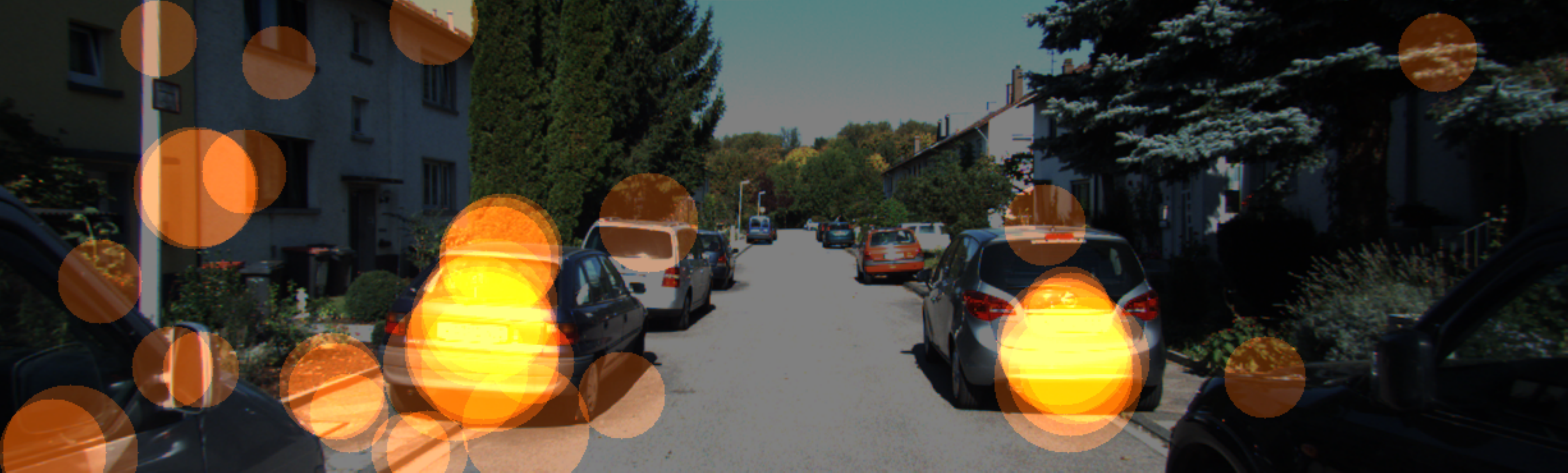}
    \caption{The circular domains based on the line features distributed in the KITTI-08 scene. The center of the circular domains is the midpoint of the corresponding line features, and the length of the line is the diameter. The color brightness shows the distribution of the line features at the local position. It is worth noting that, the corresponding circular domains of line features matched on the same object have a wide range of overlap each other.}
    \label{fig:Trck1}
\end{figure}

\begin{figure}[!t]
    \centering
    \includegraphics[width=3.4in]{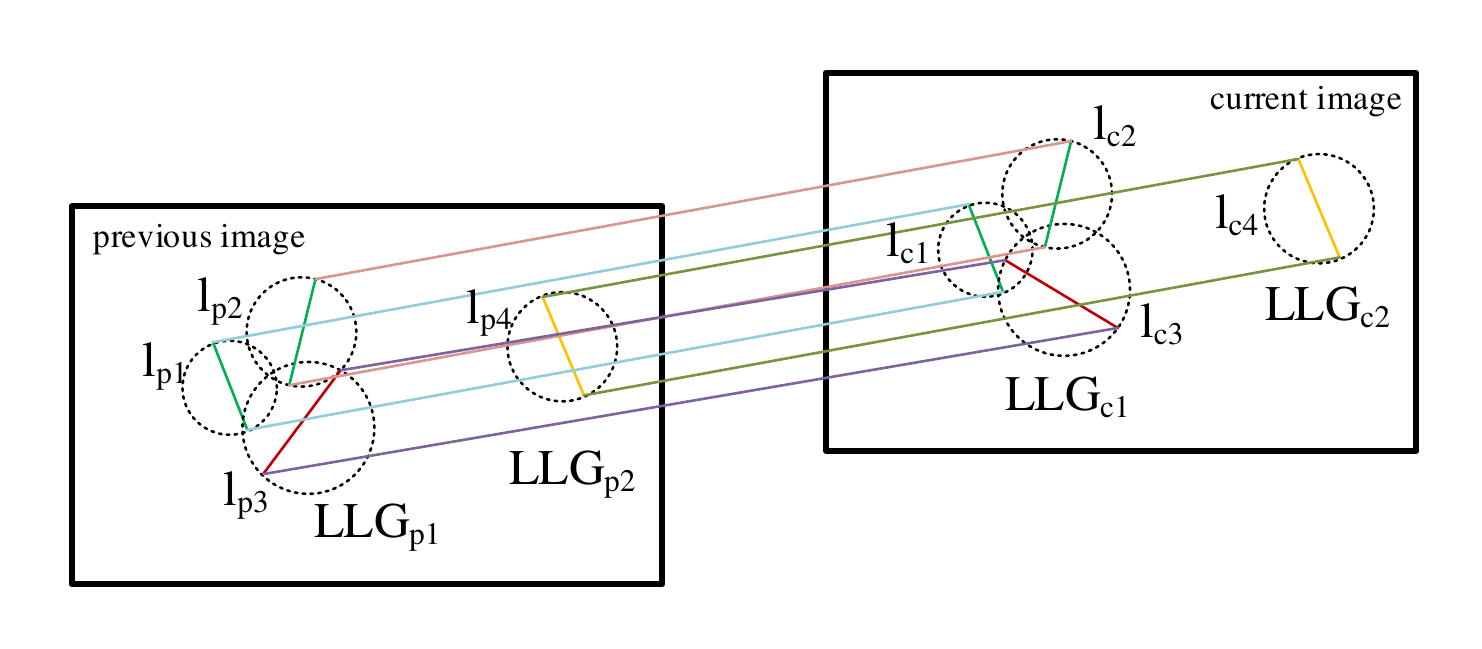}
    \caption{The line features tracking process between the previous frame and the current frame. We obtain the rough point features matching ($l_{p1,2,3,4}$ - $l_{c1,2,3,4}$) between two different frames, and they are constitute two LLG in each view, such as $LLG_{p1}$ and $LLG_{p2}$ in the previous frame. The mis-tracking line feature $l_{p3}$ - $l_{c3}$ in the $LLG_{p1}$ and $LLG_{c1}$, respectively, is obvious difference with other correct tracked lines in the same LLG. It is worth noting that the $LLG_{p2}$ which involves only one line feature, using the meaning angle and distance of all matched line feature pairs in the image as the threshold to judge the correctness of the match.}
    \label{fig:Track_2}
\end{figure}
It is naturally operation point features via the grid scheme, however, handling line segments is not a trivial task. There is no reasonable structure for enveloping line features since its large span and the different direction between lines. To overcome this problem, we focus on the straight line itself and construct the circular domain as parameterized by the midpoint and length of the line.  Figure \ref{fig:Trck1} showed the distribution of the color circular domains based on the line features in the real scene, and the color brightness respect the line features distribution in the local area. The local line features with similar characteristics have brighter colors, i.e., the corresponding circular domains of line features matched on the same object have a wide range of overlap each other. Hence, creating a local line feature domain refer to the circular domain relationship between the line features, removing the outlier matched lines based on the position and direction of all the line features in the local line domain, see figure \ref{fig:Track_2}. Like point matching above mentioned, we obtain rough matched line ($l_{p1,2,3,4}$ – $l_{c1,2,3,4}$) between the previous frame and the current frame. Specially, we determine the local line group (LLG) to line $l_{p1,2,3}$, whose circular domains overlap each other. It is noticed that the correctly tracked line  features in the sane LLG have similar positions and directions, hence, we computed the average angle and distance of mid-point between the target tracking line pair, and filter out those matches whose angle and distance are more than twice the meaning of all matched lines in the same LLG as an outlier to ensure that the correspondences are meaningful enough. It is worth mentioning that for LLG with only one line feature such as $l_{p4}$, we use the meaning angle and distance of all matched features in the image as the threshold to judge the correctness of the match.

\subsubsection{Dynamic feature detection}
\label{sec:Track_c}
\begin{figure}[!t]
    \centering
    \includegraphics[width=3.4in]{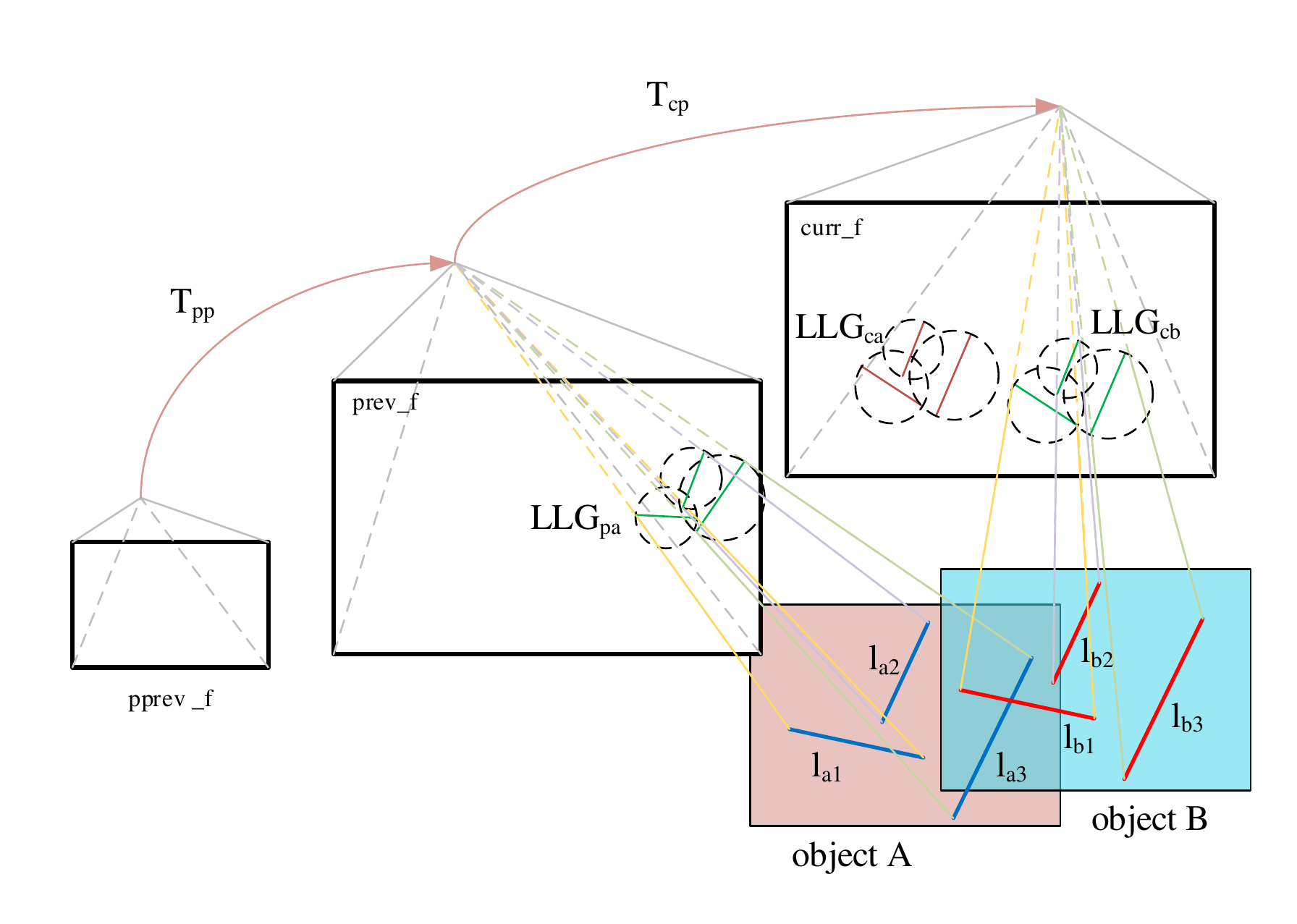}
    \caption{The dynamic line feature tracking processes between adjacent frames. $T_{pp}$ denoted the pose transformation between the previous frame $prev_f$ and its previous frame $pprev_f$, $T_{pc}$ denoted the pose transformation between the $prev_f$ and the current frame $curr_f$. We assume that the dynamic object A move to B during a sampling time, hence, the dynamic spatial line $l_{a1}, l_{a2}, l_{a3}$ on the A moved to the $l_{b1}, l_{b2}, l_{b3}$ at the same time. Noticed that there was a large error between the dynamic detected line features in the $LLG_{cb}$ and the line features in the $LLG_{ca}$ estimated by the corresponding lines in the $LLG_{pa}$ in the $prev_f$. We calculated and averaged the sum of squared line dynamic distance between the detected lines and the estimated lines in all the $LLG$s, and defined the $LLG$ as $dynamic$ $LLG$, such as $LLG_{cb}$, which dynamic value exceeded the threshold and remove the dynamic line features in the $dynamic$ $LLG$.}
    \label{fig:Track_3}
\end{figure}
After detecting and filtering the mismatching features for the current frame, we now track the matched features between the adjacent frames. However, all the tracked features contain the dynamic point and line features matched to the dynamic objects which reduced the accuracy of pose estimation.

Dynamic environment is an open challenge of visual SLAM, and addressing the issues is one of the important aspects of our work. In this step, we briefly review the dynamic point feature algorithm, $dynamic$ $grid$, which identifies dynamic point features based on the motion model. Furthermore, we introduce the dynamic line filter based on the LLG above mentioned. Normally, dynamic objects have more abnormal motion relative to the entire stationary scene, an intuitive response is that the dynamic points often have larger photometric re-projection errors between the re-projected point and the detected feature point. Finally, we also note that the feature with the same dynamic characteristics generally clusters together. Therefore, we used the motion model, i.e., the pose transformation transform between the previous frame $prev_f$ and the previous frame of the $prev_f$, to estimate the location of the matched point features in the current frame. Then, We calculated and averaged the sum of the squared Euclidean distance between the matched point features and the estimated point features of the detected point features in all the above grids. Once the value exceeds the threshold, the grid and its surrounding 8 grids are defined as the dynamic grid, and the features in the dynamic grid are dynamic. For more details about the dynamic grid, the algorithm can refer to DynPL-SVO. 

However, as we said above, the line segment with a large span and the different directions does not have a suitable structure to envelop it as the grid for the point. The above-mismatched line feature work has proved the effectiveness of the LLG method, hence, we extend the $dynamic$ $grid$ algorithm from point features to the line features in this work. As shown in figure \ref{fig:Track_3}, there is dynamic object A with severy line features ($l_{a1}, l_{a2}, l_{a3}$) whose projection constitute $LLG_{pa}$ in the previous frame, during a sample time, the object A move to the B, and the corresponding dynamic line features ($l_{b1}, l_{b2}, l_{b3}$) projected to current frame and constitute $LLG_{cb}$. In this step, we still used the motion model $T_{pp}$ of the previous frame to estimate the location of the tracked line features in the current frame, such as lines in the $LLG_{ca}$. And we defined the distance between the estimated lines in the $LLG_{ca}$ and the detected lines in the $LLG_{cb}$ as the dynamic distance of the line feature. Like the point features, we calculated and averaged the sum of the squared dynamic distance of line features in all the $LLG_{cb}$. Once the value exceeds the threshold, the $LLG_{cb}$ is defined as $dynamic$ $LLG$, and the line features in the $dynamic$ $LLG$ are the dynamic feature. After dynamic feature detection, we discard dynamic features and use the remaining static ones to estimate a more accurate camera pose for the current frame. These steps are summarized in Algorithm \ref{alg:1}.

\begin{algorithm}
    \caption{: Dynamic line detecting using $dynamic $ $LLG$}
    \begin{algorithmic}[1]
        \Require Motion model $\mathbf{T}_{pp}$ of the previous frame $prev\_f$, the matched line feature set between the current frame $curr\_f$ and $prev\_f$;
        \Ensure The $dynamic$ $LLGs$.
        \For {$ each $ $ LLG_c \in curr\_f$} 
        \For {$each $ $ l_j \in LLG_c $}
                  \State $e_{dyna}$ += $LineErr(l_j, \mathbf{T}_{pp})$
                \EndFor
                \If{$e_{dyna} > \rho$}
                  \State $LLG_c$ insert to $DynaLLGs$ 
                \EndIf
              \EndFor
            \State \Return{$DynaLLGs$}          
        \end{algorithmic}
    \label{alg:1}
\end{algorithm}
\subsubsection{Pose estimation}
\label{sec:Track_d}
We next briefly review the line features performed in the pose estimation. The point measurements are treated in a standard way as method \cite{mur2017orb}, no more details in the following. Next, we introduce a more stable line back-project model and derive its corresponding jacobian. We had obtained the line features in a sequence of images and their plural positions in camera frames, we transform the camera ego-motion estimation into a non-linear least-square equation formed by the projection constraints of the corresponding line features in the previous frame and in the current frame. Continuing the pose estimation scheme proposed in the method \cite{Ma2022DynPLSVOAN}, we introduce the horizontal re-projection error of line features into the system as additional reference information to assist in pose estimation. So the non-linear least-square equation of our approach is shown in (\ref{equ:Track_1}). For a description of the detail, the reader is referred to \cite{Ma2022DynPLSVOAN}. 

\begin{equation}
    \begin{split}
    \label{equ:Track_1}
        \xi ^* &=\underset{\xi}{arg\ min} \Bigg[ \sum_{i=1}^{m}{e_{p}^{i}(\xi)}^{T}\Sigma _{e_{p}^{i}}^{-1}e_{p}^{i}(\xi) +  \\
        &\sum_{j=1}^{n}{e_{l_v}^{j}(\xi)}^{T}\Sigma _{e_{l_v}^{j}}^{-1}e_{l_v}^{j}(\xi) +  \sum_{k=1}^{q}{e_{l_h}^{k}(\xi)}^{T}\Sigma _{e_{l_h}^{k}}^{-1}e_{l_h}^{k}(\xi) \Bigg]  
    \end{split}
\end{equation}
where $m, \,n$ and $q$ denoted the number of points, all lines, and the lines whose endpoints are not close to the edge of the image, respectively. It consisted of the point re-projection error $e_{i}^{p}$, vertical and horizontal re-projection error of line feature $e_{l_v}^{j}, e_{l_h}^{k}$. The matrices $\Sigma ^{-1}$ in (\ref{equ:Track_1}) represent the inverse covariance matrices related to the uncertainty of each re-projection error.

In this step, we force on the new representation of line features and its re-projection jacobian utilized in the motion estimation. There are various forms of 3D line representation, we assumed the line $L$ endpoints are $x_s(x1, y1, z1)$ and $x_e(x2, y2, z2)$,  respectively. Hence, the line in the word can be respected as follows: 
\begin{equation}
    L \Rightarrow \frac{x-x_1}{l} = \frac{y-y_1}{m} = \frac{z-z_1}{n}
\end{equation}
where the $l$, $m$, and $n$ refer to the $x_2-x_1$, $y_2-y_1$, and $z_2-z_1$, respectively. Furthermore, the line on the normalized plane, i.e., $z_1=1$ and $z_2=1$, can de described as:
\begin{equation}
    L \Rightarrow \frac{x-x_1}{l} = \frac{y-y_1}{m}
\end{equation}
In addition, we define the vertical re-projection errors of the line features were the distance from the endpoints of the re-projected line features $L_{pro}$ on the normalized plane to the detected line features $L_{det}$ in the current frame, shown as following:
\begin{equation}
    \begin{aligned}
        d_s = \frac{[x_s-x_1,\ y_s-y_1] \times [l,\ m]}{\sqrt{l^2 + m^2}}\\
        d_e = \frac{[x_e-x_2,\ y_e-y_2] \times [l,\ m]}{\sqrt{l^2 + m^2}}
    \end{aligned}
\end{equation}
where the $d_s$ and $d_e$ were the distance of the start point and the end point between the line features, the $x_s, y_s$ and $x_e, y_e$ denote the coordinate value of the detected line feature start point and end point on the normalized plane respectively, and the $x_1, y_1$ and $x_2, y_3$ represent the coordinate value of the projected line feature start point and end point on the normalized plane respectively. Thus, we can obtain the Jacobian of line feature vertical re-projection error using the point jacobian can be expressed as follows:
\begin{equation}
    \begin{aligned}
    \frac{\partial d}{\partial \delta \xi } & = \frac{\frac{[x_s-x_1,\ y_s-y_1] \times [l,\ m]}{\sqrt{l^2 + m^2}}}{\partial \delta \xi}\\
    & = \frac{\partial (m\,e_s - l\,e_e)}{\partial \delta \xi}\frac{1}{\sqrt{l^2 + m^2}}\\
    & = \begin{bmatrix}m & -l\end{bmatrix}
    \frac{\partial \begin{bmatrix}e_s \\ e_e\end{bmatrix}}{\partial \delta \xi}
    \frac{1}{\sqrt{l^2 + m^2}}
    \end{aligned}   
\end{equation}
where the $e_s$ and $e_e$ represent the re-projection error between the line features start point and end point, respectively.

Finally, we obtain the incremental motion estimation between the two consecutive frames, which can be modeled by the non-linear least-square equation (\ref{equ:Track_1}) with jacobian of all features.

\subsubsection{Gloable gray similar (GGS) algorithm}
\label{sec:Track_e}
\begin{figure}
	\centering
	
	\subfloat[]{
	    \subfloat{
		\begin{minipage}[!t]{0.83in}
			\centering
			\includegraphics[width=0.85in]{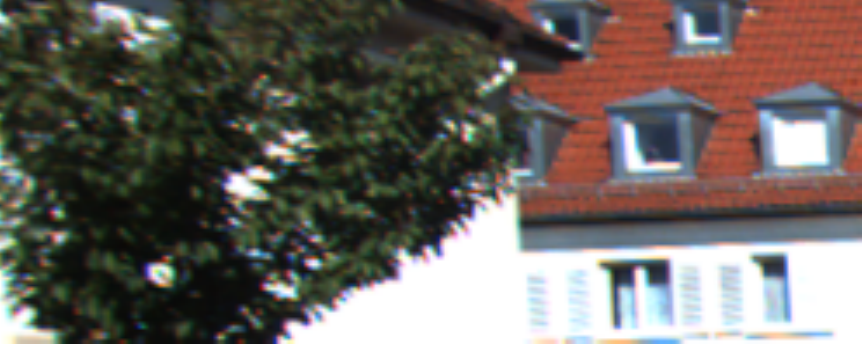}\\
			\vspace{0.05cm}
			\includegraphics[width=0.85in]{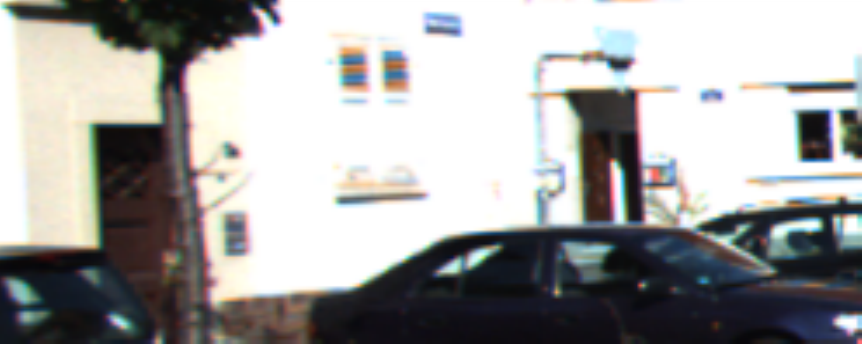}\\
			\vspace{0.05cm}
			\includegraphics[width=0.85in]{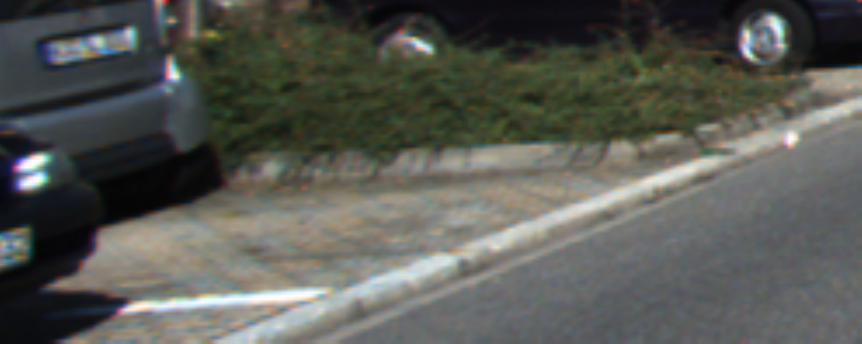}\\
			\vspace{0.05cm}
		\end{minipage}%
	    }%
	    \subfloat{
		    \begin{minipage}[!t]{0.83in}
			\centering
			\includegraphics[width=0.85in]{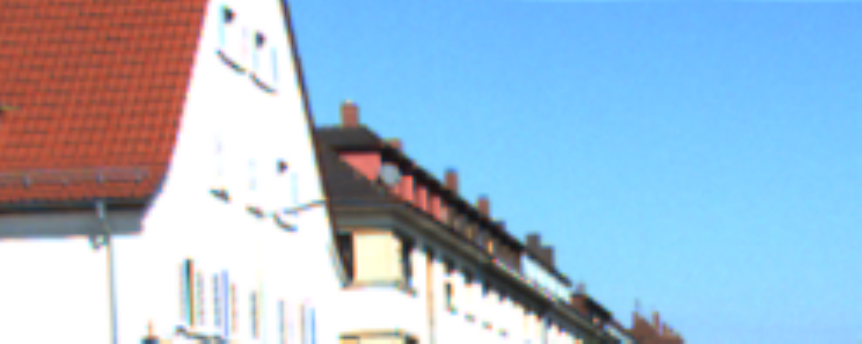}\\
			\vspace{0.05cm}
			\includegraphics[width=0.85in]{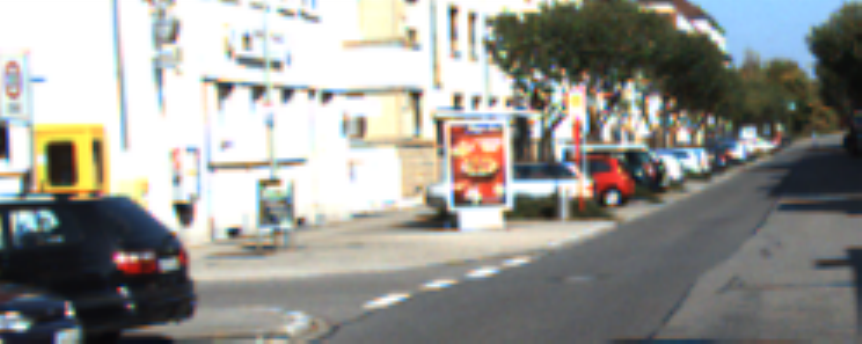}\\
			\vspace{0.05cm}
			\includegraphics[width=0.85in]{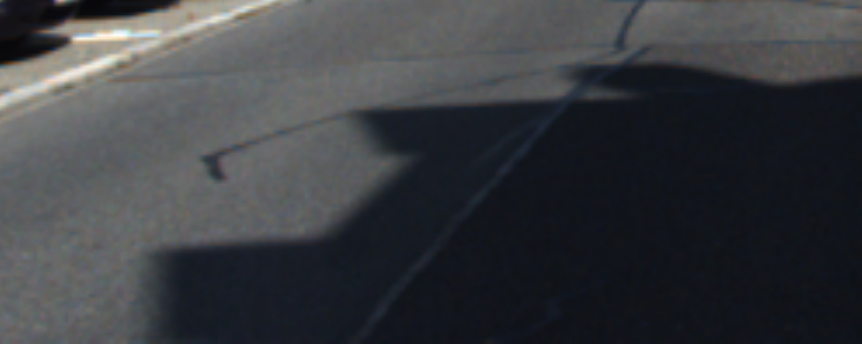}\\
			\vspace{0.05cm}
		\end{minipage}%
	    }%
	    \subfloat{
		    \begin{minipage}[1t]{0.83in}
			\centering
			\includegraphics[width=0.85in]{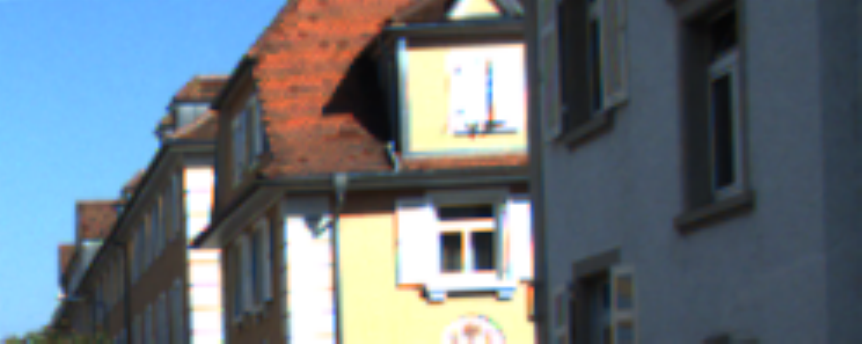}\\
			\vspace{0.05cm}
			\includegraphics[width=0.85in]{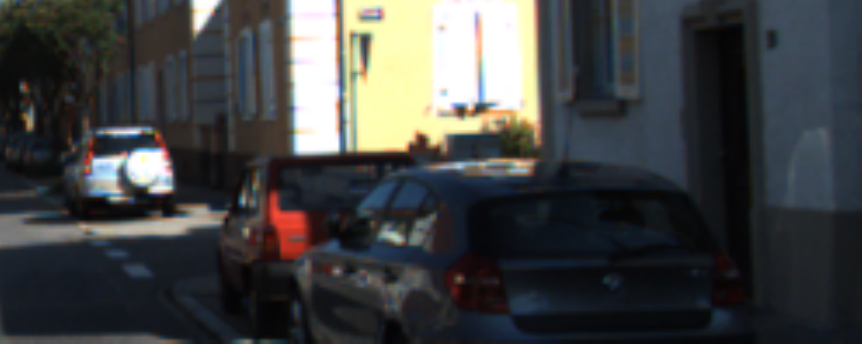}\\
			\vspace{0.05cm}
			\includegraphics[width=0.85in]{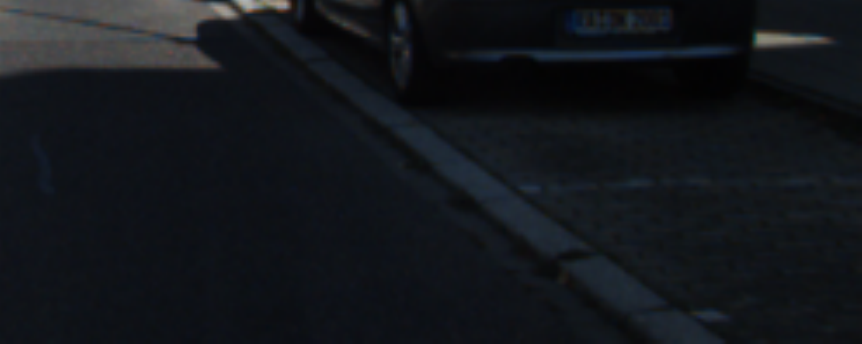}\\
			\vspace{0.05cm}
		    \end{minipage}%
	    }%
	    \subfloat{
		    \begin{minipage}[!t]{0.83in}
			\centering
			\includegraphics[width=0.85in]{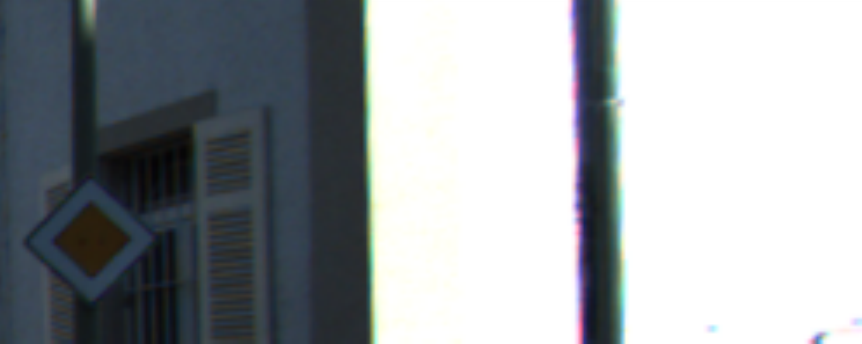}\\
			\vspace{0.05cm}
			\includegraphics[width=0.85in]{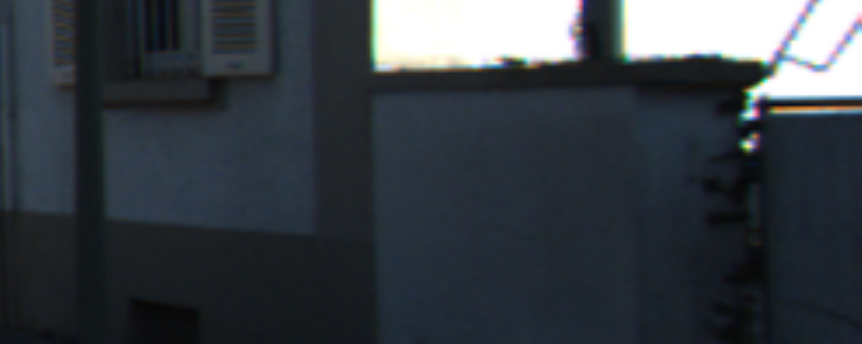}\\
			\vspace{0.05cm}
			\includegraphics[width=0.85in]{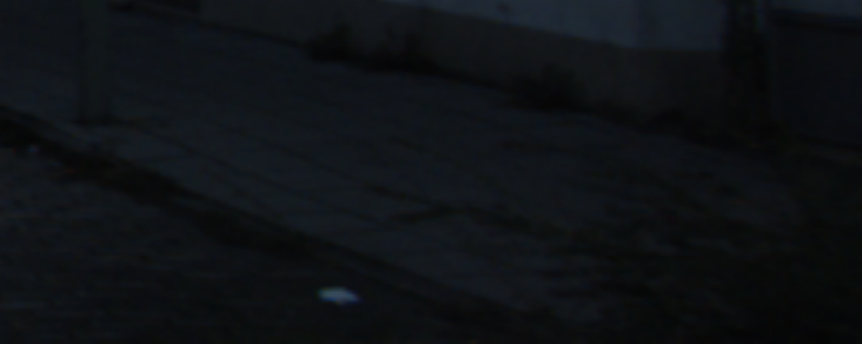}\\
			\vspace{0.05cm}
		    \end{minipage}%
	    }
	    \setcounter{subfigure}{1}
	}
	
	\subfloat[]{
	    \subfloat{
		\begin{minipage}[!t]{0.83in}
			\centering
			\includegraphics[width=0.85in]{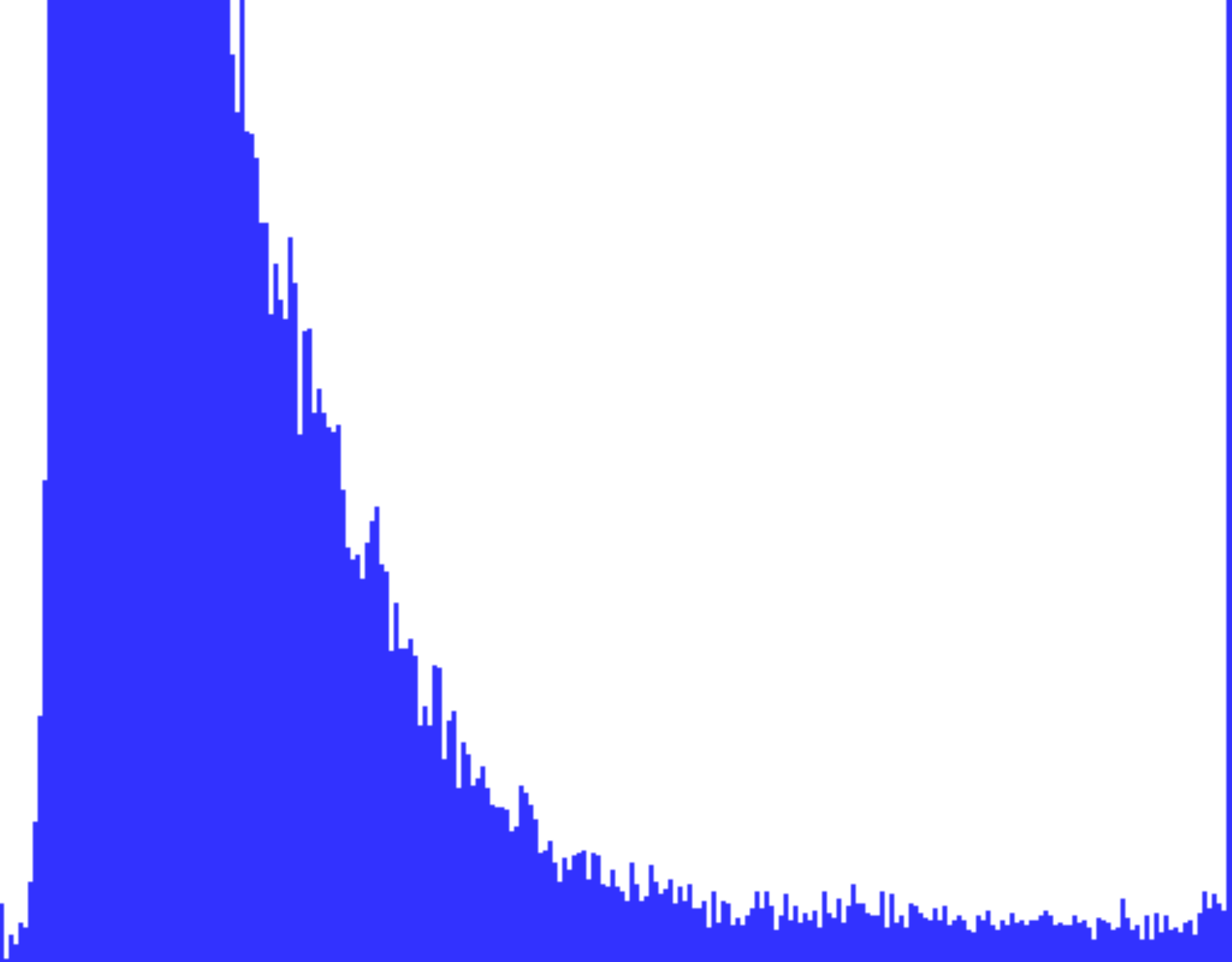}\\
			\vspace{0.05cm}
			\includegraphics[width=0.85in]{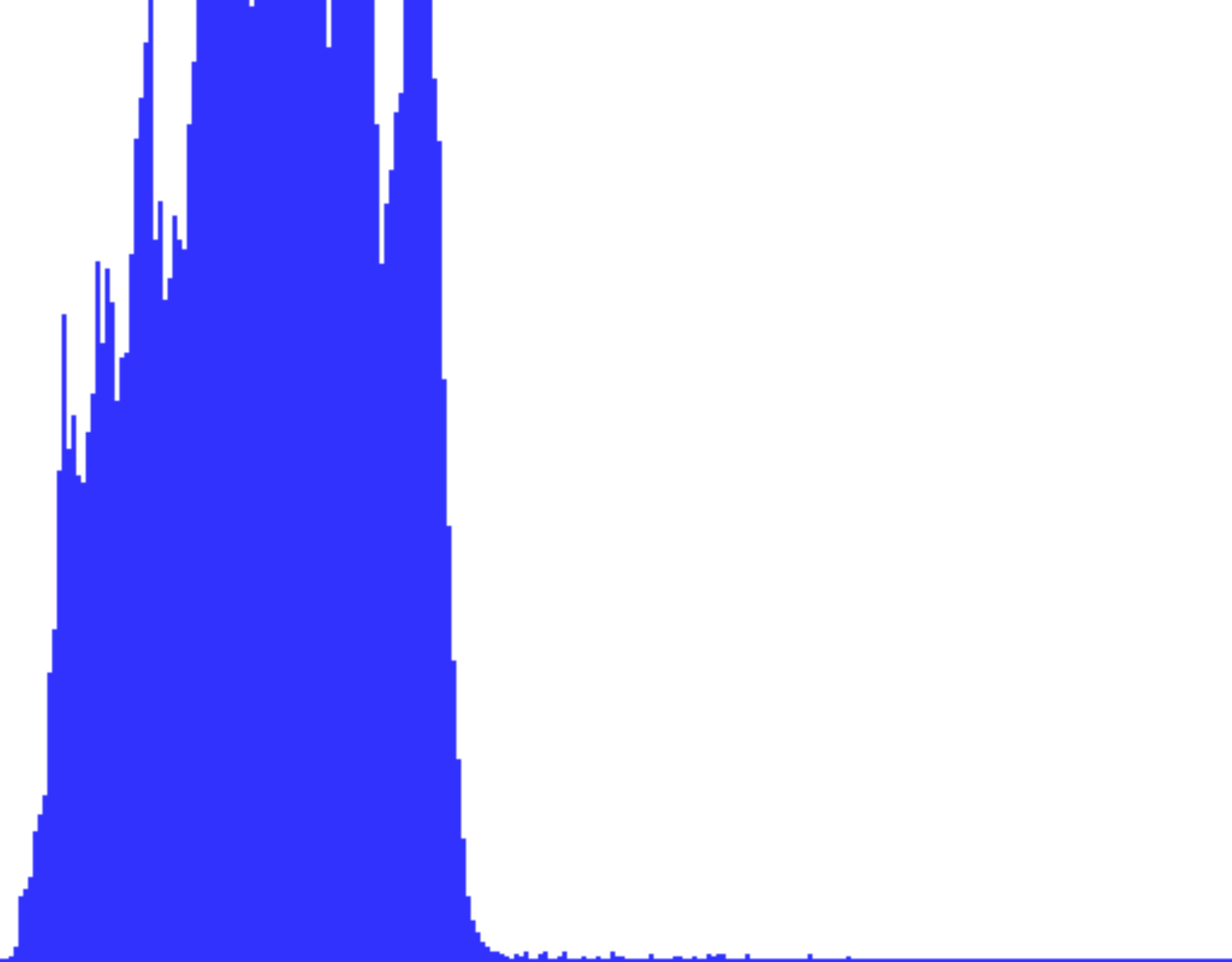}\\
			\vspace{0.05cm}
			\includegraphics[width=0.85in]{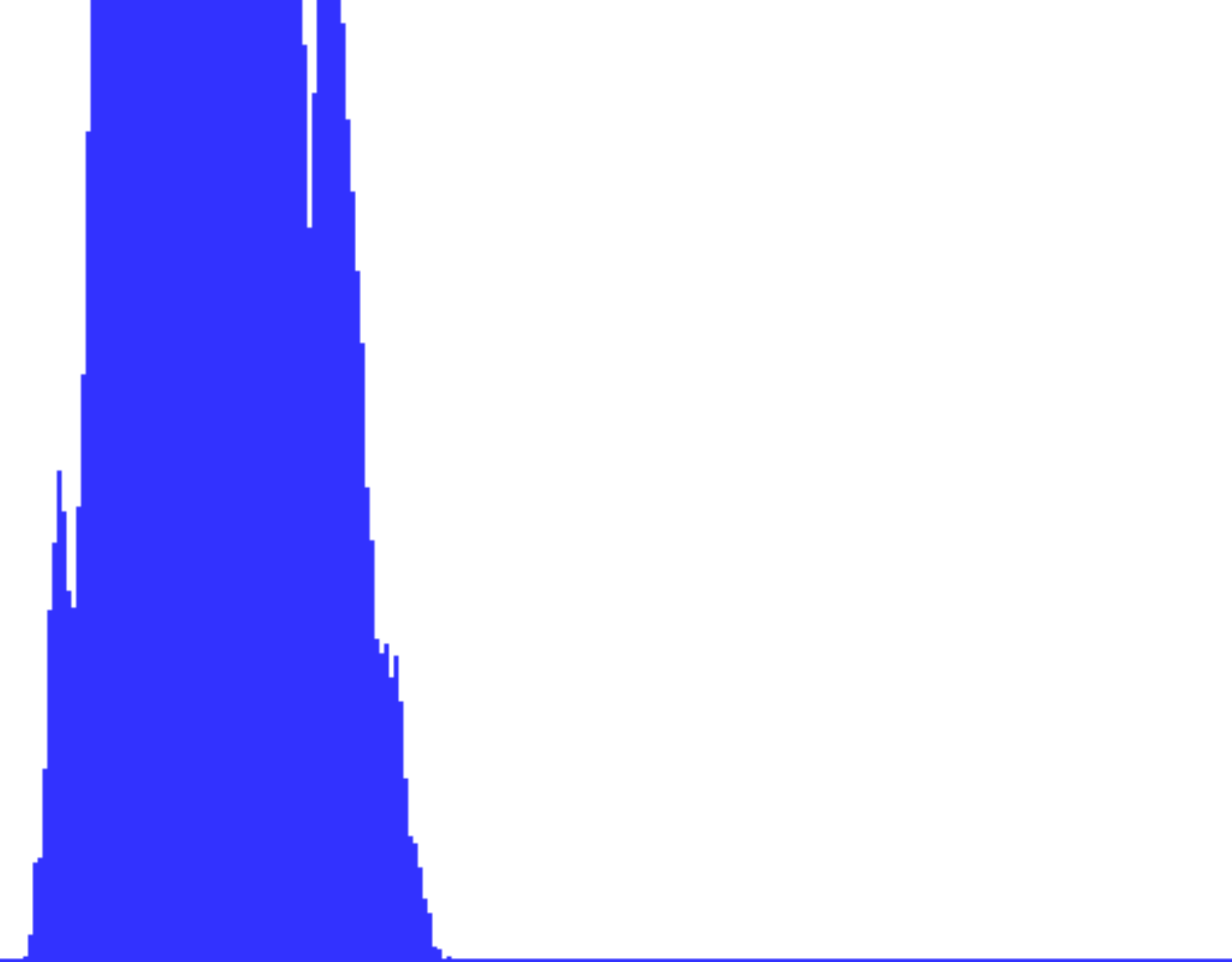}\\
			\vspace{0.05cm}
		\end{minipage}%
	    }%
	    \subfloat{
		    \begin{minipage}[!t]{0.83in}
			\centering
			\includegraphics[width=0.85in]{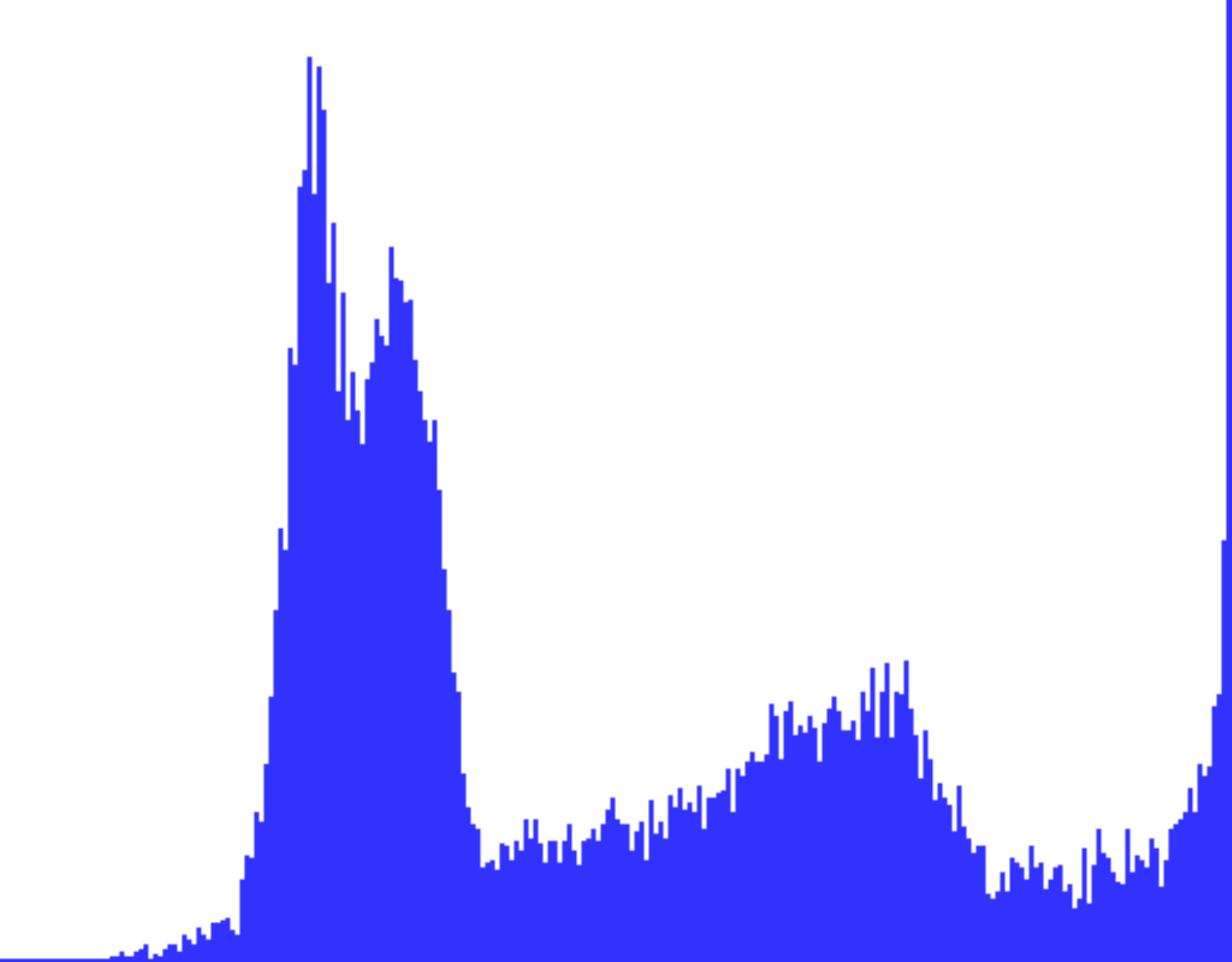}\\
			\vspace{0.1cm}
			\includegraphics[width=0.85in]{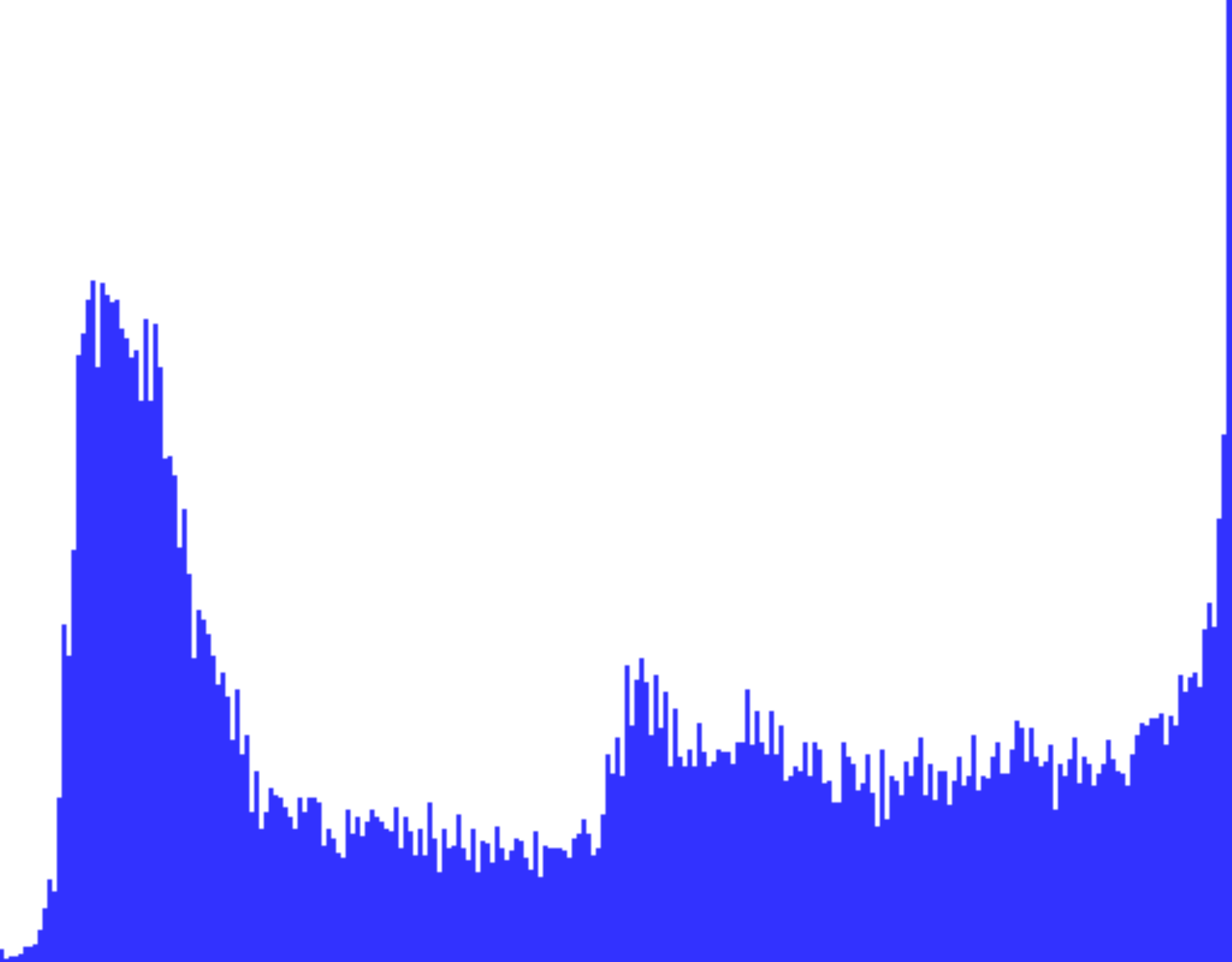}\\
			\vspace{0.1cm}
			\includegraphics[width=0.85in]{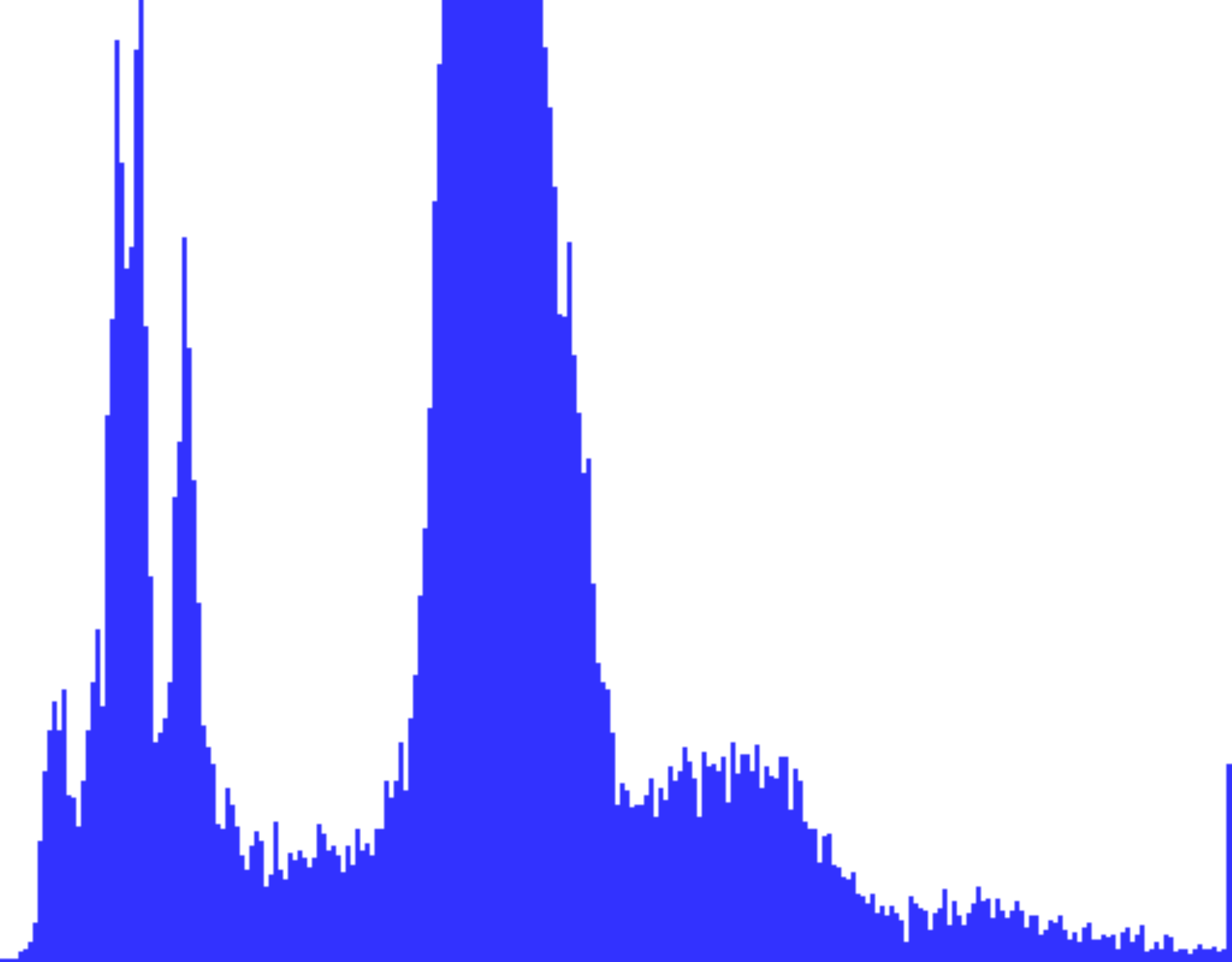}\\
			\vspace{0.1cm}
		\end{minipage}%
	    }%
	    \subfloat{
		    \begin{minipage}[1t]{0.83in}
			\centering
			\includegraphics[width=0.85in]{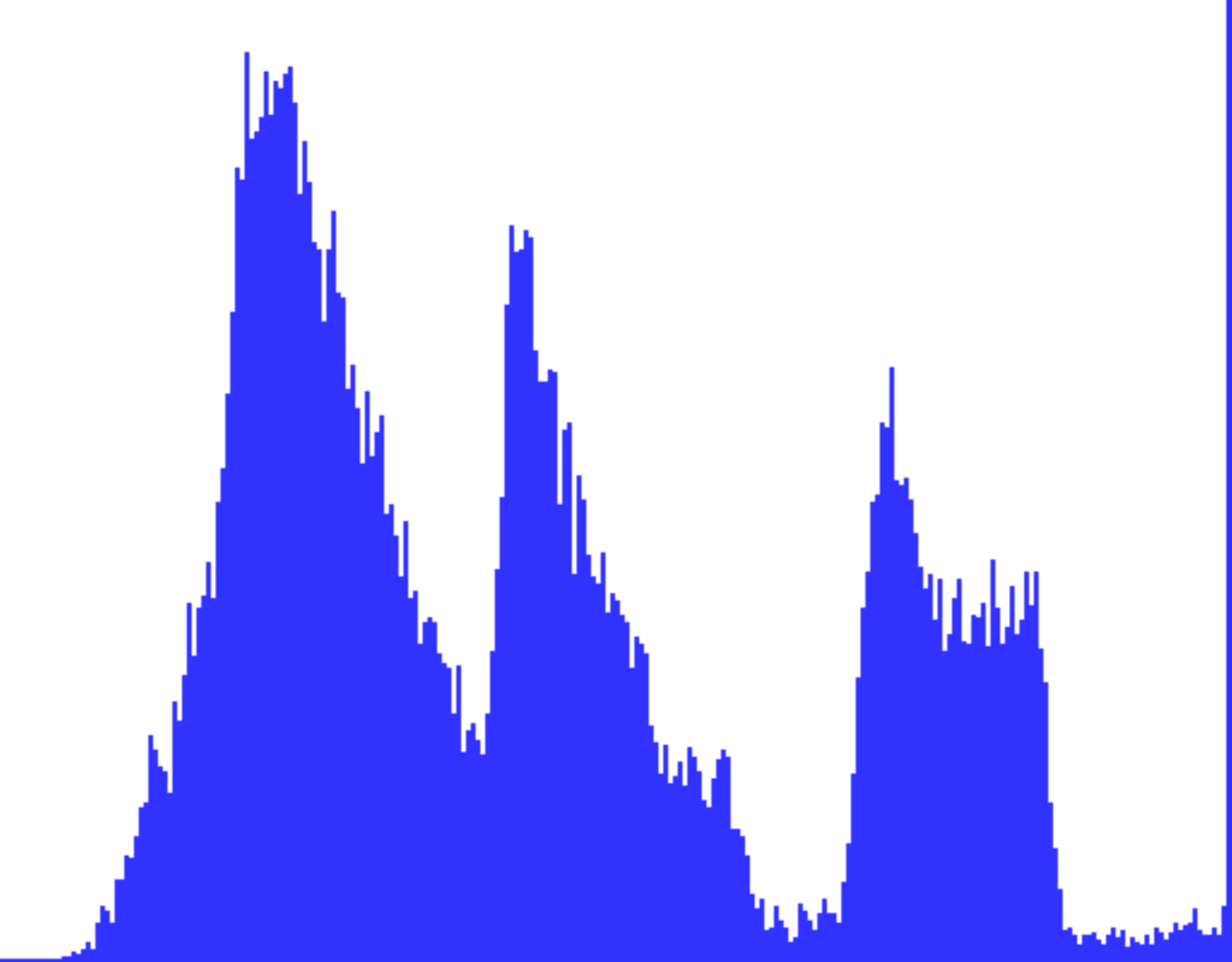}\\
			\vspace{0.1cm}
			\includegraphics[width=0.85in]{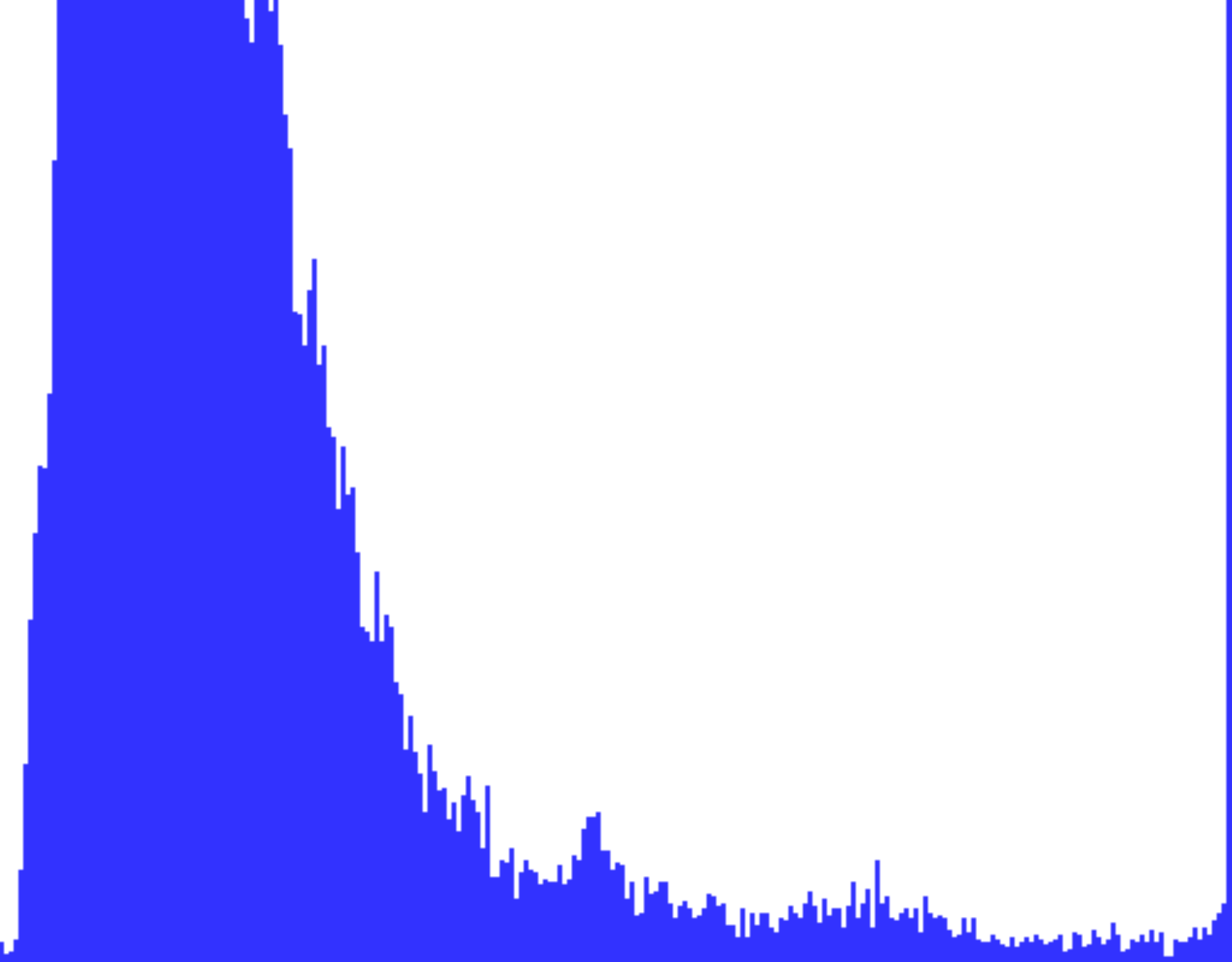}\\
			\vspace{0.1cm}
			\includegraphics[width=0.85in]{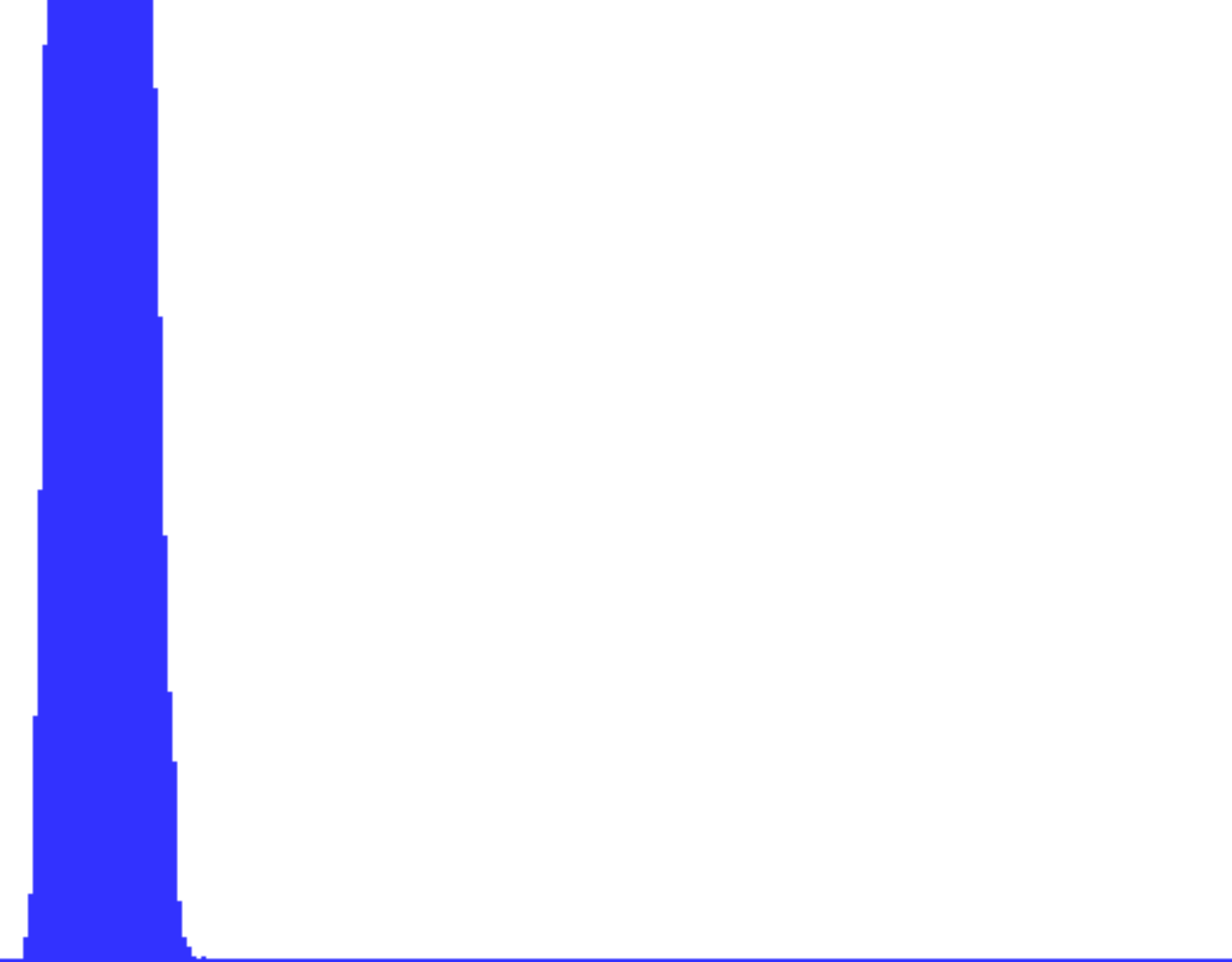}\\
			\vspace{0.1cm}
		    \end{minipage}%
	    }%
	    \subfloat{
		    \begin{minipage}[!t]{0.83in}
			\centering
			\includegraphics[width=0.85in]{images/Technical_Approach/GGS/3.pdf}\\
			\vspace{0.1cm}
			\includegraphics[width=0.85in]{images/Technical_Approach/GGS/7.pdf}\\
			\vspace{0.1cm}
			\includegraphics[width=0.85in]{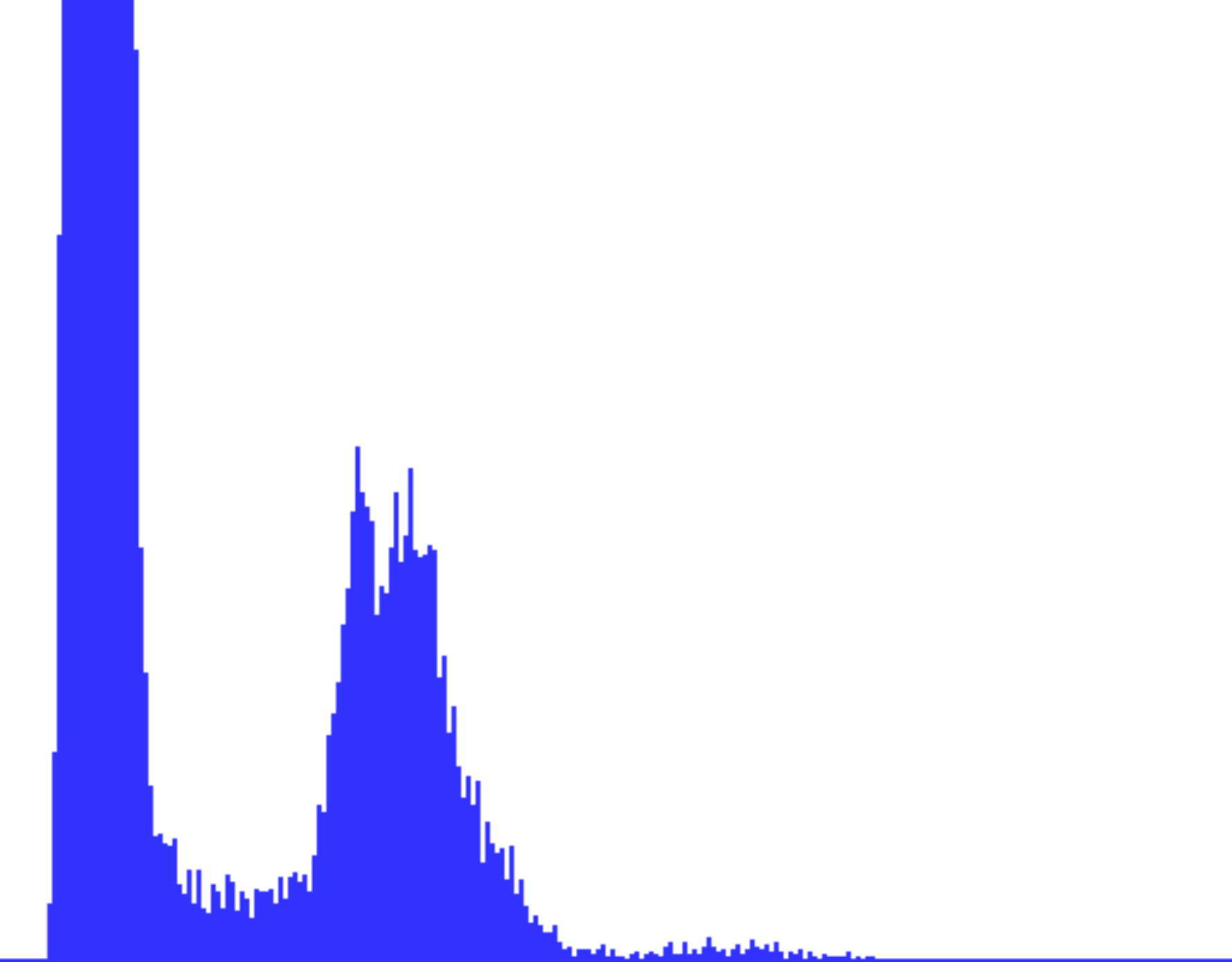}\\
			\vspace{0.1cm}
		    \end{minipage}%
	    }
	    \setcounter{subfigure}{2}
	}
	
	\subfloat[]{
	    \subfloat{
		\begin{minipage}[!t]{0.83in}
			\centering
			\includegraphics[width=0.85in]{images/Technical_Approach/GGS/0.pdf}\\
			\vspace{0.05cm}
			\includegraphics[width=0.85in]{images/Technical_Approach/GGS/3.pdf}\\
			\vspace{0.05cm}
			\includegraphics[width=0.85in]{images/Technical_Approach/GGS/7.pdf}\\
			\vspace{0.05cm}
		\end{minipage}%
	    }%
	    \subfloat{
		    \begin{minipage}[!t]{0.83in}
			\centering
			\includegraphics[width=0.85in]{images/Technical_Approach/GGS/1.pdf}\\
			\vspace{0.1cm}
			\includegraphics[width=0.85in]{images/Technical_Approach/GGS/5.pdf}\\
			\vspace{0.1cm}
			\includegraphics[width=0.85in]{images/Technical_Approach/GGS/9.pdf}\\
			\vspace{0.1cm}
		\end{minipage}%
	    }%
	    \subfloat{
		    \begin{minipage}[1t]{0.83in}
			\centering
			\includegraphics[width=0.85in]{images/Technical_Approach/GGS/2.pdf}\\
			\vspace{0.1cm}
			\includegraphics[width=0.85in]{images/Technical_Approach/GGS/6.pdf}\\
			\vspace{0.1cm}
			\includegraphics[width=0.85in]{images/Technical_Approach/GGS/10.pdf}\\
			\vspace{0.1cm}
		    \end{minipage}%
	    }%
	    \subfloat{
		    \begin{minipage}[!t]{0.83in}
			\centering
			\includegraphics[width=0.85in]{images/Technical_Approach/GGS/3.pdf}\\
			\vspace{0.1cm}
			\includegraphics[width=0.85in]{images/Technical_Approach/GGS/7.pdf}\\
			\vspace{0.1cm}
			\includegraphics[width=0.85in]{images/Technical_Approach/GGS/11.pdf}\\
			\vspace{0.1cm}
		    \end{minipage}%
	    }
	    \setcounter{subfigure}{3}
	}
	\centering
	\caption{}
	\vspace{-0.2cm}
	\label{fig:Track_4}
\end{figure}

Compared with the classical feature-based algorithm, such as ORB\_SLAM2, our method utilize the pixel-level intensity as image similarity evaluation parameters instead of features and designed the keyframe selection and the loop closure detection based on global gray similar (GGS). For every input frame, in order to improve the real-time of the system, the GGS computing, and the feature extraction for left and right are in parallel. The GGS computing steps are given as shown in figure \ref{fig:Track_4}. First, We grayed the left image of the new frame and performed scaling with the scaling factor (set to 0.8 in our experiments). Then, we evenly divide the original gray-scale image and the scaled image into 3$\times$4 grids (see figure \ref{fig:Track_4} (a)) and save the pixel intensity distribution in each grid separately, i.e., each frame GGS can be respected as 24 integer arrays with a size of 256 shown as figure \ref{fig:Track_4} (b), (c). It is worth noting that there is work in the system based on the GGS of the frame, such as keyframe selection and loop closure detection.

\subsection{Local mapping}
\label{sec:LM}
The local mapping thread is activated when the frame is selected as the keyframe. In this section, we describes the behavior of the local mapping which can be divide two subsection: $local$ $map$ $updata$ and $local$ $map$ $BA$.

\subsubsection{Local map updata}
The local map consists of keyframes, 3D landmarks both points and lines, and the co-visibility graph which involve the nodes, i.e., the keyframe poses, and the edges between keyframes when sharing a minimum number of landmarks (set to 20 in our experiments). Thus, the local map update only when a new keyframe is selected and inserted. In this work, we present a time-saving keyframe selection based on the GGS of the frames and the effectiveness of the strategy is verified by the effect of the system. Since the similarity between local frames decreases with sample time and the adjacent frames with high similarity. Thus, we calculation the GGS threshold $GGS_{th}$ through the following expression:
\begin{equation}
    GGS_{th} = GGS_{pKF} + 0.4 * (GGS_{pKF+1} - GGS_{ppKF+1})
\end{equation}
where $GGS_{pKF}$ and $GGS_{pKF+1}$ represent the GGS of the previous frame $prev\_f$ and the GGS of its first consecutive frame, and the $GGS_{ppKF+1}$ refer to the GGS of the previous frame of the $prev\_f$. If the GGS of current frame exceed the $GGS_{th}$, then the current frame is inserted into the system as a new KF, updating the new keyframe GGS and waiting for the next keyframe selected.

Ones the new keyframe was detected, we insert the keyframe into the back-end of the system and update the local map to optimate the element state in the local map. First, we build data association between the current KF with local KFs and obtain a consistent set of common features observed in them. Furthermore, for the new common features observed in the current KF and previous KF, creating the corresponding 3D landmark and insert them into the local map.

\subsubsection{Local BA}
To eliminate the drift of element states in the local map, such as KFs transformation, 3D point pose, and the 3D line landmark length and direction in the world coordinate. After the KF insertion, the next step is to build the co-visibility graph model and perform a local BA to optimize the element in the local map. First, we need to build data association between the KFs which connect with newly inserted keyframes that share at least 20 landmarks, and common features 
observed between the KEs in the co-visibility graph. These will be used as elements to build optimization models and perform BA which minimizes the projection errors between the observations and the landmarks projected to the frames where they were observed. The local BA pipelines are similar to the ones presented in the PL-SLAM. It should also be underlined that we follow a similar approach to the one described in section \ref{sec:Track_d} as it introduced line horizontal re-projection error into the BA framework.

\subsection{Loop closure detection}
\label{sec:LC}
The loop closure detection aims to reduce the trajectory estimation drift and improve the accuracy of building maps. The essence of loop detection is to identify the similarity between the field of view and optimize the pose graph in the loop. It is important to remark that, in this paper, we propose an accurate and fast loop closure detection approach based on the GGS of inserted keyframes, which are not needed to build huge offline or online feature dictionaries from different image datasets. In the following, we first introduced detecting loop closures from the GGS approach and then described the correction of the pose estimations of the KFs involved in the loop.

\subsubsection{Loop detection}
\label{sec:LC_1}

\begin{figure}[!t]
    \centering
    \subfloat[]{\includegraphics[width=3.4in]{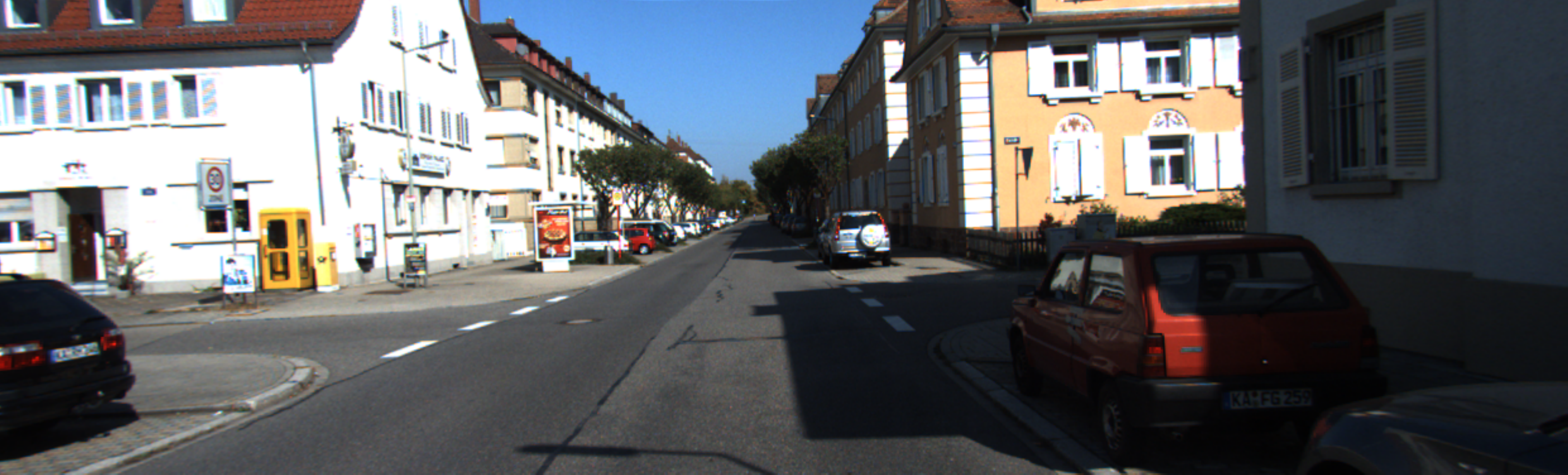}}
    
    \subfloat[]{\includegraphics[width=3.4in]{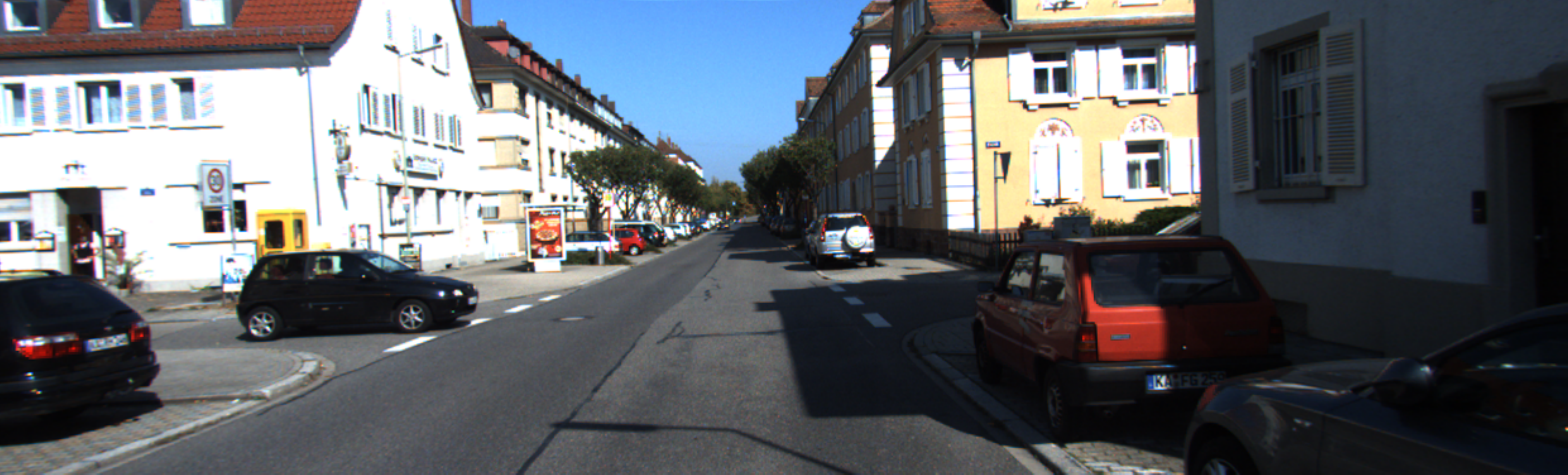}}
    \caption{
    The pictures of the corrcet loop closure position in the KITTI-00. comparing with the historical view (a), the later (b) emerged a car in the left of the view, which seriously reduce the LCD performance.}
    \label{fig:LC_1}
\end{figure}

For each input keyframe, we computed its GGS in section \ref{sec:Track_e} and record the similarity with historical KFs. Compared to building the huge BoW model for point and line features in the frame and computing image similar to all historical keyframes by dictionary model, the loop closure detection based on the GGS has an SOTA performance with BoW and better real-time. No matter the GGS-based or Bow-based LCD approachs, there are two requirements to be met in the candidate detection step, first, toleranting the position of the image to be identified, i.e., the images near the correct loop position can be detected as a candidate. The second, robustness to the dynamic scenes, i.e., when the system is located in the correct loop position and there is a new object exited in the view compared with the last time such as cars, which seriously reduce the LCD performance (see figure \ref{fig:LC_1}). Considering the above challenges, we define the similarity, $sim_v$ between two images by GGS as follows:

\begin{equation}
    \begin{aligned}
    sim_v = \frac{1}{M * N}\sum_{i=1}^{12} \min \left \| hist_a(p, \, i) - hist_b(q,\, i)\right \|,\\ p, \,q=\{0, \, 1\}
    \end{aligned}
\end{equation}
Where $M,\, N$ denote the size of the image and the $hist(0, \, i)$ and $hist(1, \, i)$ represent the ith grid gray distribution histogram value of the original and scaled image, respectively. 

It is noticed that the similar evaluation based on the GGS through the scaled scheme to meet the LCD requirement tolerant to the identity image, and the grid strategy in the GGS computign process improves the LCD performance in the dynamic scenes. To avoid the greatest negative impact of incorrect loop detection on the system, we further filter the candidates with the following tests. First, the ratio of matched feature and the transformation between the current keyframe and looped keyframe are important informations for checking the consistency of the loop closure candidate. Hence, we merged the feature matching results and the transformation information into a scalar, through the following expression:
 \begin{equation}
     lc_{rat} = \alpha \left \| \log(\mathbf{T}_i\mathbf{T}_j^{-1}) \right \| * ratio_{inl}
 \end{equation}
 where the $\mathbf{T}_i, \,\mathbf{T}_j$ represent the current keyframe and candidate looped keyframe transformation, respectively, and the $ratio_{inl}$ was the initial inlier features ratio (was set 50\% in our experimentation, and the $\alpha$ in the expression was set 0.001). If the value of $lc_{rat}$ lies below some pre-established threshold, which in our experiments has been set to 20\%, then the candidate was the outlier. Second, generally, adjacent view images have similar characteristics, including loop detection characteristics. Thus, we assume that the preview keyframe has significant similarity with the looped keyframe, and there is a significant similarity between the current keyframe and the adjacent frames of the looped keyframe. The above conditions can ensure the correctness and efficiency of the LCD.

\subsubsection{Loop correction}
\label{sec:LC_2}
After obtained the correct loop closure, we computed the transformation between the current keyframe and looped keyframe and optimized the pose of all the KFs in the loop. This is a typically graph optimization problem, which can be solved by formulating the problem as a PGO. Establishing the corresponding graph model whose node is the pose of the keyframe and the edges are the data association between keyframes which has at least 20\% commonly features. In this paper, we solved the problem using the g2o library \cite{grisetti2011g2o} as a method PL-SLAM. Updating the pose of keyframes in the loop and optimizing the state of the landmarks observed by these keyframes to correct the map information. Finally, when the system is finished, we implemented the global BA to optimize all the keyframe pose and global map landmark state.

\section{Experimental Results and Analysis}
In this section, we evaluated the proposed method on the public KITTI dataset [33] and EuRoC MAV dataset [34]. To verify the performance of our method, we compared the accuracy of our algorithm with the state-of-the-art VO systems that can run with stereo RGB data. The selected methods were stvo pl [27], ORB SLAM2 [6] and BoPLW [35]. In
order to fairly compare all methods, we only kept the front end of ORB SLAM2 and BoPLW to form VO systems. In addition, to illustrate the benefits of the line feature horizontal re-projection error on the accuracy of motion estimation, we showed the results of our proposal with only horizontal or vertical re-projection errors. Finally, we evaluated the ability of dynamic grids to cope with dynamic scenes on highly dynamic sequences, where the trajectory of a stereo camera in several video sequences with moving objects was estimated.

All the experiments were tested on an Intel Core i5-4210U CPU @ 1.70GHz × 4 and 16 GB RAM without GPU. Since there are no dynamic objects in the EuRoC MAV dataset, we only estimated the trajectory of our method with or without dynamic grids in the KITTI dataset.

\subsection{Dynamic Objects Detection Experiments}

\begin{figure*}[!t]  
    \centering    
    \subfloat[] 
    {
        \begin{minipage}[!t]{0.45\textwidth}
            \centering          
            \includegraphics[width=\textwidth]{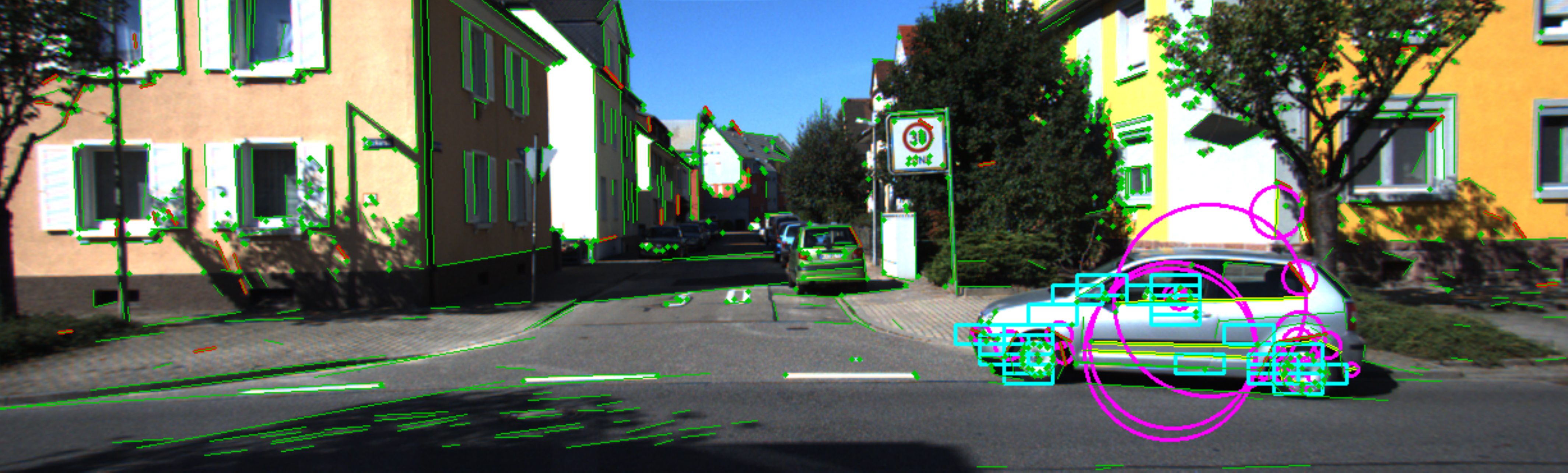}   
        \end{minipage}%
    }
    \subfloat[] 
    {
        \begin{minipage}[!t]{0.45\textwidth}
            \centering      
            \includegraphics[width=1\textwidth]{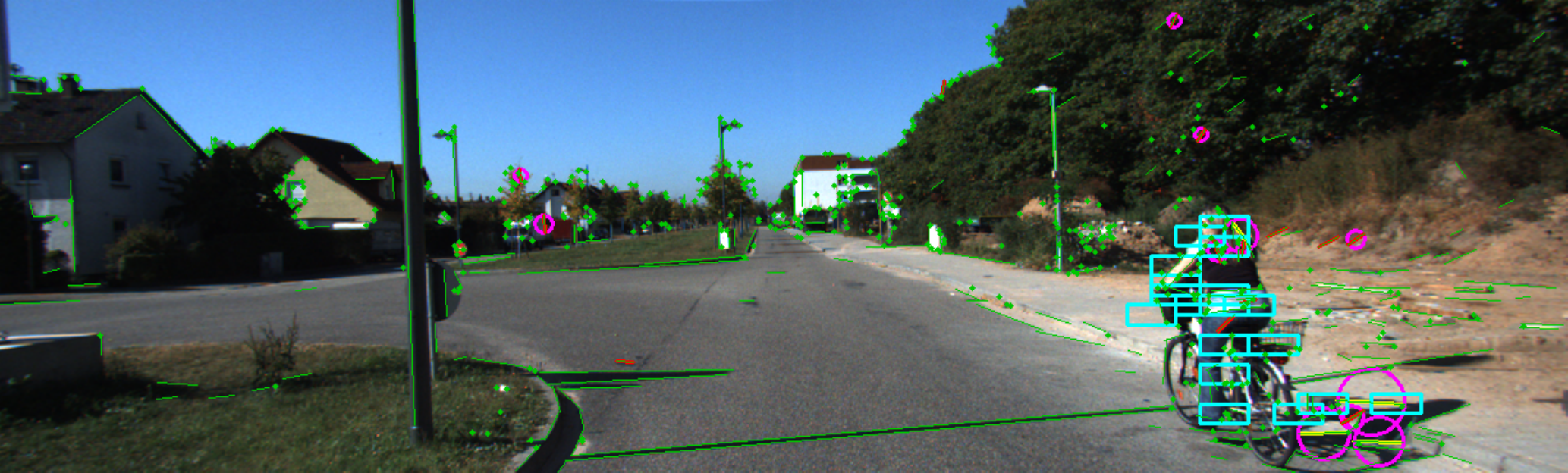}   
        \end{minipage}
    }
    
    \subfloat[] 
    {
        \begin{minipage}[!t]{0.45\textwidth}
            \centering      
            \includegraphics[width=1\textwidth]{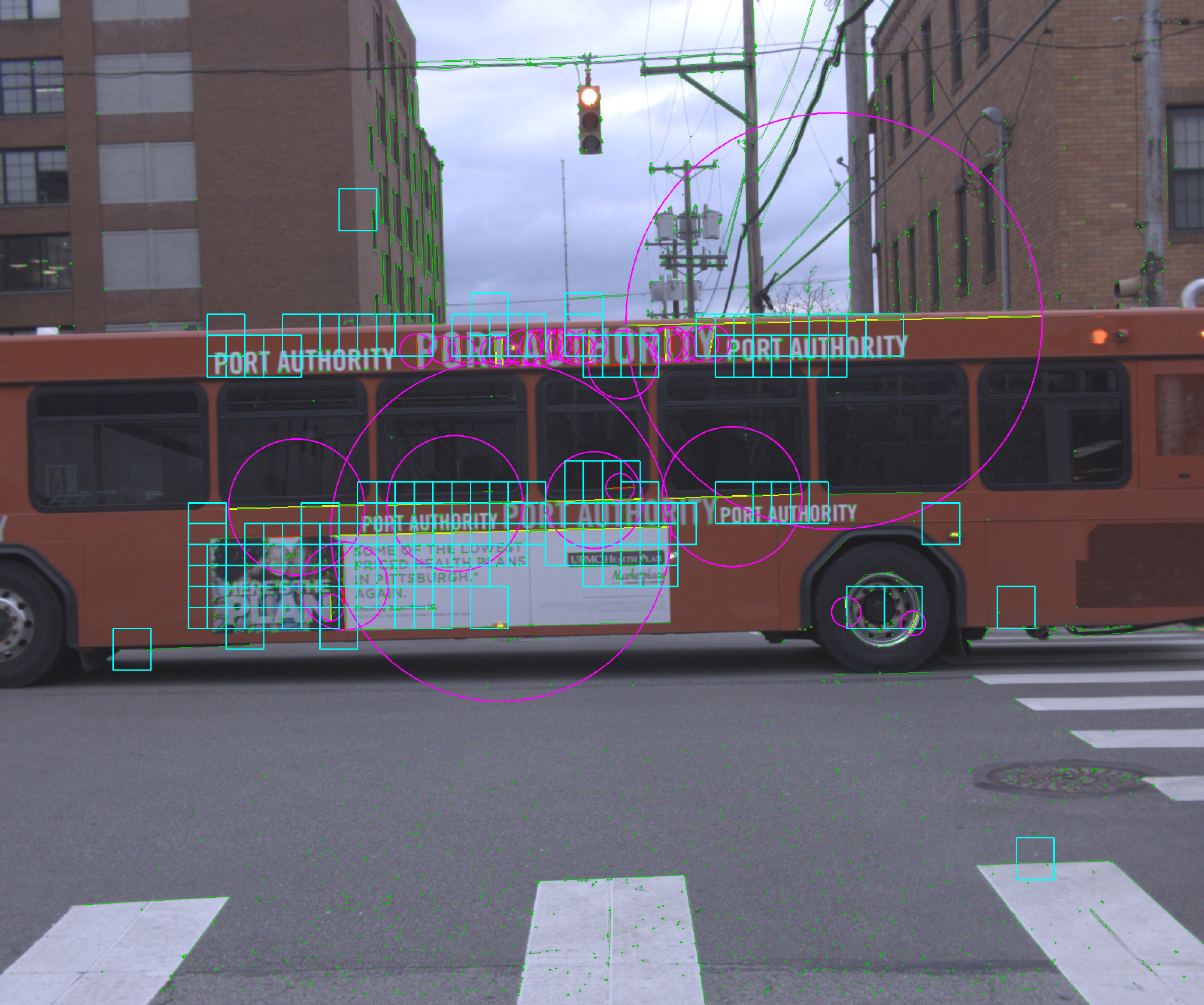}   
        \end{minipage}
    }
    \subfloat[] 
    {
        \begin{minipage}[!t]{0.45\textwidth}
            \centering      
            \includegraphics[width=1\textwidth]{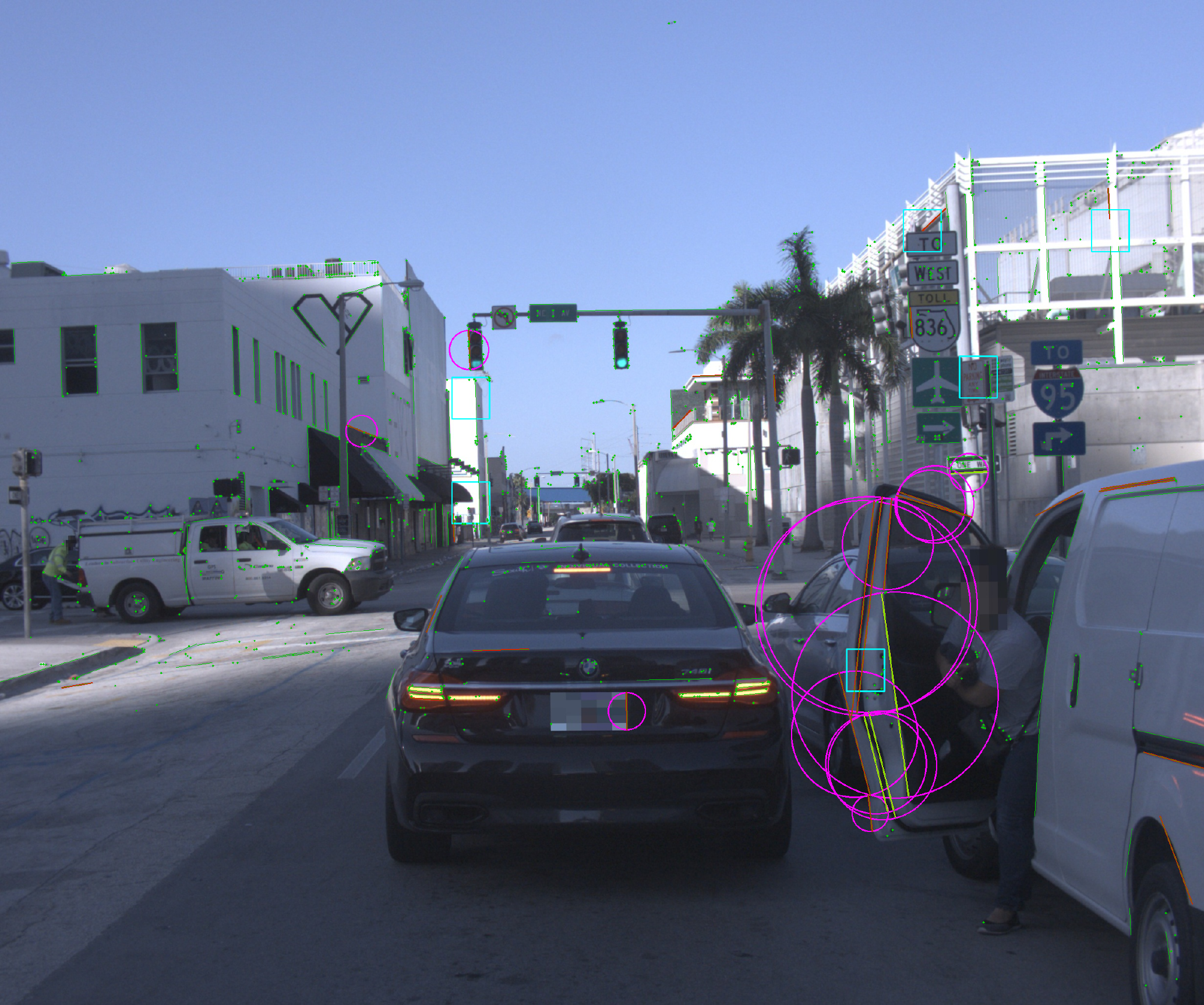}   
        \end{minipage}
    }
    \caption{Dynamic scenes on the \textit{KITTI} and \textit{Argoverse 1} dataset. (a) Moving cars in \textit{KITTI-07}. (b) The person riding a bike in \textit{KITTI-06}. (c) (d) Challenging dynamic scene during automatic driving in the \textit{Argoverse 1} dataset} 
    \label{fig:exp1}  
\end{figure*}

\begin{table*}[!t]
\centering
 \caption{Mean Absolute RMSE errors on the KITTI Dataset.}
 
 \begin{tabular}{c | c  c | c  c | c  c | c  c} 
    \multicolumn{1}{c|}{}&\multicolumn{2}{c|}{ wo/dynamic grid \& LLG}&\multicolumn{2}{c|}{ w/dynamic grid}&\multicolumn{2}{c|}{ w/dynamic LLG}&\multicolumn{2}{c}{ w/dynamic grid \& LLG}\\
    Seq. & t & R & t & R & t & R & t & R\\
 \hline\hline
00 & 4.414051 & 1.141609 & 4.1586 & 1.036655 & 4.467157 & 1.189108 & \textbf{4.098689} &\textbf{1.027319} \\
01 & 69.670172 & \textbf{4.016659} & 64.208694 & 5.589265 & 69.674846 & 4.020314 & \textbf{64.200222} & 5.589637 \\
02 & 9.858701 & 2.0002 & \textbf{9.097316} & \textbf{1.776921} & 9.797926 & 1.976967 & 18.140115 & 3.742419 \\	
03 & 5.776864 & 4.315284 & 5.807239 & 4.157304 & \textbf{5.77685} & 4.315294	& 5.807236 & \textbf{4.157297} \\
04 & 2.622336 & 49.376319 & \textbf{2.253605} & \textbf{43.416013} & 2.622333 & 49.376311	& 2.25378 & 43.427358 \\
05 & 2.069462 & 0.676671 & 2.021802	& 0.717747 & 2.060681 & \textbf{0.663488} & \textbf{2.014489} & 0.705174 \\	
06 & 4.577771 & 3.468494 & \textbf{4.537722} & 3.543528 & 4.558604 & \textbf{3.40538} & 4.538055 & 3.49156 \\	
07 & 1.207638 & 0.615184 & 1.109332 & 0.57397 & 1.207379 &0.614653 & \textbf{1.104154} & \textbf{0.573476} \\
08 & 5.647390 & 1.940569 & 4.723587 & 1.512542 & 5.625841 & 1.938448 & \textbf{4.707446} & \textbf{1.508939} \\
09 & 4.447918 & 1.731804 & \textbf{4.126671} & \textbf{1.412788} & 4.447922 & 1.731822 & 4.127799 & 1.413146\\	
10 & 2.067542 & 1.2582 & \textbf{1.975722} & 1.127534 & 2.067256 & 1.258117 & 1.976234 & \textbf{1.127434}\\
\end{tabular}
\label{tab:exp1}
\end{table*}

\subsubsection{KITTI Dataset}
\subsubsection{Argoverse 1 Dataset}
Dealing with dynamic scenes is the one of major connect of this paper. The dynamic scenes approach in the PLD-SLAM include two parts : the dynamic grid and dynamic LLG. The dynamic grid follows \cite{Ma2022DynPLSVOAN}, and the latter is used to detect the dynamic line features. To evaluate the ability of the system to cope with dynamic scenes, we test the proposed method on well-known dataset KITTI and the Argoverse 1 dataset \cite{chang2019argoverse}, which have 7 high-resolution ring cameras (1920 × 1200) recording at 30 Hz with overlapped field of view providing 360◦ coverage. In addition, there are 2 front-facing stereo cameras (2056×2464) sampled at 5 Hz. The dynamic scenes containing moving objects like cars and pedestrians in some image sequences, showed as in Fiurexxx have a great impact on the performance of the VO/vSLAM systems. The dynamic object detection verified that the dynamic grids and dynamic LLG can accurately identify the dynamic regions to avoid the influence of dynamic features on the accuracy. 

To further represent the error in the quantitative analysis, we compared the absolute and relative root-mean-square error (RMSE) with and without dynamic grid approaches in the KITTI dataset,
as shown in table xxx.

\subsection{Loop Closure Detection Experiments}
\subsubsection{KITTI Dataset}
\subsubsection{NEU Dataset}

To evaluate the proposed loop closure detection, we test its performance on the KITTI dataset and the high similarity scenes indoor stereo sequences namely $NEU$ $Dataset$, we have record using MYNT@EYE 120° stereo camera. In addition, to further explane the performance the proposed LCD, we compare the LCD with GGS to STOA LCD approach based on BoW model and openMAP, which ... and the result explane in the table xxx and Figure xxx showed 

\subsection{Localization Accuracy and Efficenty Experiments}
\subsubsection{KITTI Dataset}
\subsubsection{EuRoC MAV Dataset}
\subsubsection{Argoverse 1 Dataset}
\subsubsection{NEU Dataset}

We evaluate the performance of PLD-SLAM in terms of the localization accuracy and computation time. To comprehensively evaluate the implementation effect of the method, we used the public dataset KITTI, EuRoC dataset, in addition, the short term daatset Argoverse 1 and the high similiarity  dataset collected by us to be used in the experimental analysis.

\subsubsection{KITTI Dataset} The KITTI dataset


\balance
\bibliographystyle{ieeetr}
\bibliography{Trans_on_Robotics}
\end{document}